\documentclass{article}

\usepackage{makecell}
\usepackage[final]{lg}

\usepackage[utf8]{inputenc} 
\usepackage[T1]{fontenc}    
\usepackage{hyperref}       
\usepackage{url}            
\usepackage{booktabs}       
\usepackage{rotating}
\usepackage{amsfonts}       
\usepackage{amsmath}
\usepackage{nicefrac}       
\usepackage{microtype}      
\usepackage[dvipsnames]{xcolor}
\usepackage{kotex}
\usepackage{adjustbox}
\usepackage{multirow}
\usepackage{amssymb}
\usepackage{longtable}
\usepackage{comment}
\usepackage[autostyle]{csquotes}
\usepackage{hhline}
\usepackage{tabu}
\usepackage{tablefootnote}
\usepackage{longtable}
\usepackage{graphicx}
\usepackage{array}
\usepackage{caption}
\usepackage{tcolorbox}
\usepackage{lipsum}
\usepackage{relsize}
\usepackage{colortbl}
\usepackage{tablefootnote}
\tcbuselibrary{breakable}
\usepackage{floatpag}
\floatpagestyle{plain}
\usepackage{float}
\usepackage{threeparttable}
\usepackage{enumitem}
\setlist[enumerate,itemize]{leftmargin=1.5em}

\newcommand{\model}{EXAONE 4.0 }
\newcommand{\comp}{LG~AI~Research}

\title{\centering EXAONE 4.0: Unified Large Language Models Integrating Non-reasoning and Reasoning Modes}
\author{%
  \comp\thanks{The complete list of authors who contributed to this work can be found in Appendix~\ref{appendix:contributors}.
}\\
}

\begin{document}

\maketitle
\addtocounter{footnote}{-1}

\begin{abstract}

This technical report introduces EXAONE 4.0, which integrates a \textsc{Non-reasoning} mode and a \textsc{Reasoning} mode to achieve both the excellent usability of EXAONE 3.5 and the advanced reasoning abilities of EXAONE Deep. To pave the way for the agentic AI era, EXAONE 4.0 incorporates essential features such as agentic tool use, and its multilingual capabilities are extended to support Spanish in addition to English and Korean. The EXAONE 4.0 model series consists of two sizes: a mid-size 32B model optimized for high performance, and a small-size 1.2B model designed for on-device applications. The EXAONE 4.0 demonstrates superior performance compared to open-weight models in its class and remains competitive even against frontier-class models. The models are publicly available for research purposes and can be easily downloaded via \url{https://huggingface.co/LGAI-EXAONE}.

\end{abstract}

\section{Introduction}

As part of LG AI Research's EXAONE foundation model series, the EXAONE language models have been developed to support diverse real-world applications through strong instruction-following and reasoning capabilities. 

The previous version, EXAONE 3.5~\citep{research2024exaone35serieslarge}, focused on real-world usability by strengthening comprehensive instruction-following abilities, while EXAONE Deep~\citep{research2025exaonedeepreasoningenhanced} emphasized reasoning performance, particularly in mathematical and coding domains. 

With the upcoming era of agentic AI in mind, EXAONE 4.0 introduces agentic tool use---a core capability for this paradigm---and further advances reasoning abilities. 

In terms of tool use, the model is developed to enable the integration of various external tools to develop agents or applications. Regarding reasoning performance, the capabilities of EXAONE 4.0 have been improved by leveraging the validated methodologies developed in EXAONE Deep. 
Notably, EXAONE 4.0 unifies both \textsc{Non-reasoning} mode---enabling rapid thinking and responses---and \textsc{Reasoning} mode---designed for deep thinking and more accurate answers---into a single model, allowing users to experience both modes within one model. 

Compared to previous versions of EXAONE, the number of tokens used during pretraining is significantly increased to bolster world knowledge. To further enhancement of expert knowledge, curating training data from specialized domains such as STEM (Science, Technology, Engineering, and Mathematics) fields plays important role on downstream tasks. Furthermore, the extension of maximum context length of the model to 128K tokens enables handling of various tasks based on significantly longer contexts, thereby improving usability. One notable challenge in processing long contexts is the computational burden of attention calculations. To mitigate this, a hybrid architecture combining global attention and local attention is adopted. This approach minimizes performance degradation while reducing computational costs during training and inference.

Moreover, the EXAONE 4.0 officially add Spanish to its supported languages, expanding beyond its previous bilingual support for English and Korean. The development of Spanish language support was designed to minimize any negative impact on English and Korean performance while maintaining the same tokenizer and vocabulary as the previous EXAONE 3.5 and Deep models.

EXAONE 4.0 particularly excels in areas focused on world knowledge and reasoning, especially in mathematical and coding domains. Despite integrating \textsc{Non-reasoning} mode and \textsc{Reasoning} mode, it secures competitive performance in instruction following. The model also shows commendable performance in long context tasks, particularly excelling in document QA (Question Answering) and RAG (Retrieval Augmented Generation) tasks frequently used by real users. Regarding tool use, it achieves a level comparable to competing models, marking the beginning of the fundamental capabilities essential for the upcoming agentic AI era. Additionally, our supported languages now include English, Korean, and the newly supported Spanish. The EXAONE 4.0 model demonstrates competitive performance in both Korean and Spanish across a diverse range of tasks, including expert-level knowledge and mathematical reasoning.

\section{Modeling}

\subsection{Model Configurations}

\begin{figure}[t!]
    \centering
    \includegraphics[width=\textwidth]{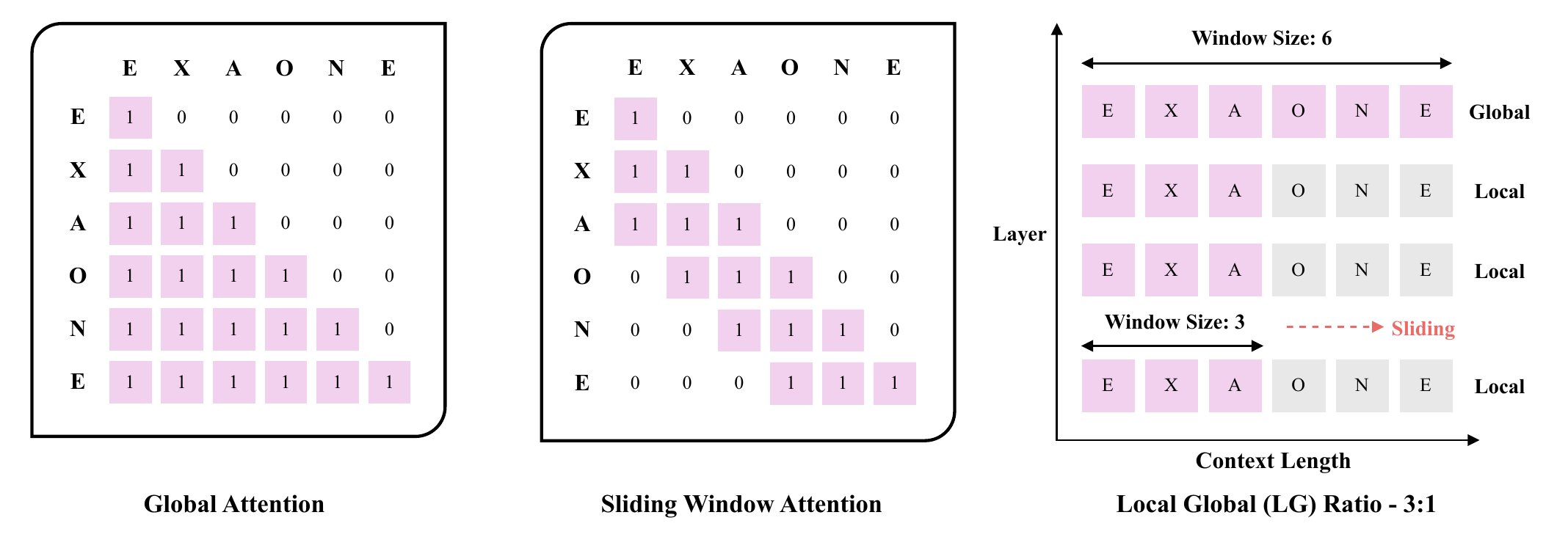}
    \caption{Visualization of the hybrid attention mechanism when the window size for local attention (sliding window attention) is set to 3. This figure illustrates how context tokens are processed across layers under the hybrid attention mechanism, highlighting the interaction between local and global attention.}
    \label{fig:hybrid_attention}
\end{figure}

The EXAONE 4.0 model retains a similar structural framework to the EXAONE 3.5 model, but incorporates several key differences in its architecture. Notably, we modify the approach to the attention mechanism. In the previous EXAONE 3.5 model, every layer utilized global attention, whereas the EXAONE 4.0 model employs a hybrid attention mechanism that combines local attention (sliding window attention type) with global attention in a 3:1 ratio, as illustrated in Figure~\ref{fig:hybrid_attention}.

Contrary to past findings that models using global attention across all layers performed better, recent studies~\citep{gemma2,gemma3,llama4} suggested that utilizing a larger window size (e.g., from 512 to 1,024 or 4,096) and applying global attention to only a minority of layers can still achieve excellent long-context performance. Additionally, it reported that incorporating small amounts of global attention periodically in combinations with heterogeneous structures like Mamba~\citep{nvidia2025nemotronhfamilyaccurateefficient,lieber2024jambahybridtransformermambalanguage,minimax2025minimax01scalingfoundationmodels} helped maintain the ability to understand global context.

In designing the EXAONE 4.0 model, a sliding window size of 4K is selected to minimize any adverse effects on short-context performance. Furthermore, the model does not employ Rotary Position embedding~\citep{yang2025ropenopeagainnew} for global attention, ensuring that the model does not develop biases towards length and can maintain a global view. For the design of the local attention mechanism, we do not employ the chunked attention strategy. Instead, we adopt sliding window attention, a well‑established form of sparse attention that offers strong theoretical stability. Unlike chunked attention, sliding window attention benefits from wide support in open‑source frameworks, ensuring robust implementation and ease of integration. To prevent performance degradation in short-context areas during long context fine-tuning, the EXAONE 4.0 model employs a careful data selection methodology and a progressive training recipe, effectively balancing efficiency and performance.

\begin{figure}[t!]
    \centering
    \includegraphics[width=\textwidth]{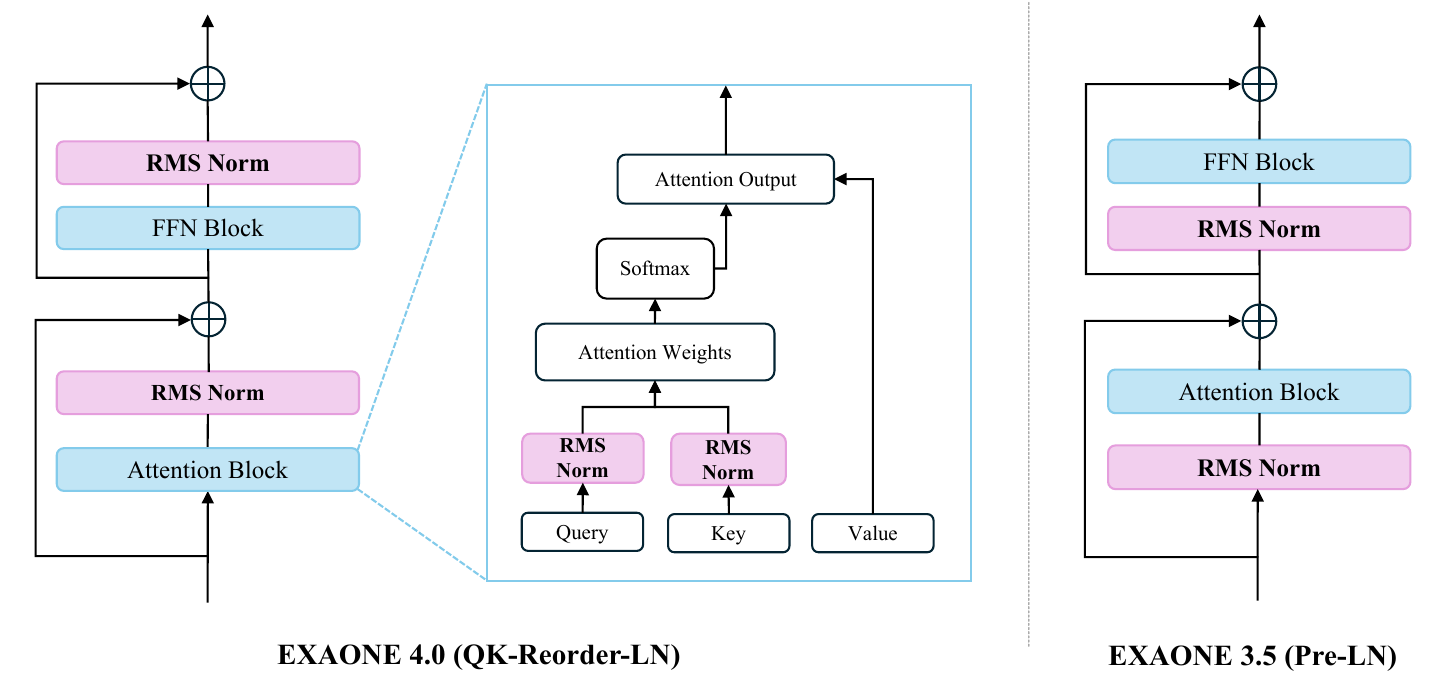}
    \caption{Visualization of repositioning layer normalization. The LayerNorm is applied after input queries and keys, and it is performed after attention output again. The type of normalization is RMSNorm.}
    \label{fig:layer_norm}
\end{figure}

Another significant change in the EXAONE 4.0 model is the repositioning of layer normalization (LayerNorm), as shown in Figure~\ref{fig:layer_norm}. According to recent studies \citep{sun2025cursedepthlargelanguage}, some layers that do not significantly impact model performance are found mainly in deep layers. This issue is attributed to the Pre-LN transformer architecture~\citep{xiong2020layernormalizationtransformerarchitecture}, which enhances stability but leads to exponentially increasing variance in outputs as model depth increases. A simple operation to control variance by providing more scaling to outputs as layers deepen was proposed, but we find that the QK-Reorder-LN method~\citep{olmo20252olmo2furious,muennighoff2025olmoeopenmixtureofexpertslanguage}, which applies LayerNorm after input queries and keys and performs LayerNorm after attention output, yields better performance on downstream tasks despite consuming more computation. The normalization type RMSNorm, which was applied since EXAONE 3.0, is retained in EXAONE 4.0.

Finally, the EXAONE 4.0 model series consists of two configurations: 32B and 1.2B. These models share the same vocabulary, which consists primarily of Korean and English tokens in roughly equal proportions, along with a tiny number of multilingual tokens, as detailed in Table~\ref{tab:mod_cfg}.

\begin{table}[b]
    \centering
    \small
    \setlength{\doublerulesep}{1pt}
    \begin{tabular}{l|cc}
        \toprule
        Model size & 32B & 1.2B \\
        \midrule
        $d$\_model & 5,120 & 2,048  \\
        Number of layers & 64 & 30 \\
        Normalization & QK-Reorder-LN & QK-Reorder-LN\\
        \midrule
        Non-linearity & SwiGLU \citep{shazeer2020gluvariantsimprovetransformer} & SwiGLU \\
        Feedforward dimension & 27,392 & 4,096 \\
        \midrule
        Attention type & Hybrid & Global \\
        Head type & GQA \citep{ainslie2023gqatraininggeneralizedmultiquery} & GQA \\
        Number of heads & 40 & 32 \\
        Number of KV heads & 8 & 8 \\
        Head size & 128 & 64 \\
        Max sequence length & 131,072 & 65,536\\
        RoPE theta \citep{su2023roformerenhancedtransformerrotary} & 1,000,000 & 1,000,000 \\
        \midrule
        Tokenizer & BBPE \citep{wang2020bbpe} & BBPE \\
        Vocab size & 102,400 & 102,400 \\
        Tied word embedding & False & True \\
        \midrule
        Knowledge cut-off & Nov. 2024 & Nov. 2024 \\
        \bottomrule
    \end{tabular}
    \vspace{2mm}    
    \caption{Configurations of \model{}language models. Key differences from previous versions include a hybrid attention mechanism and modified normalization.}
    \label{tab:mod_cfg}
\end{table}

\subsection{Pre-training}

The amount of data and computational resources used for pretraining in the EXAONE 4.0 models are summarized in Table~\ref{tab:pretraining}. For the EXAONE 3.5 32B model, 6.5 trillion tokens are used for pretraining. In comparison, the EXAONE 4.0 32B model doubles this amount, utilizing 14 trillion tokens for pretraining. This increase in data is specifically aimed at enhancing the model's world knowledge. As will be discussed later, this approach yields noticeable improvements in benchmarks that rely on knowledge, such as MMLU-Redux~\citep{gema2025mmlu}, where the use of more extensive training data has a demonstrable impact on performance.

\begin{table}[t!]
    \centering
    \small
    \setlength{\doublerulesep}{1pt}
    \begin{tabular}{l|cc}
        \toprule
        Model size & 32B & 1.2B \\
        \midrule
        Size of pretraining data (tokens) & 14T & 12T \\
        Amount of computation (FLOPs) & $2.69 \times 10^{24}$ & $8.65 \times 10^{22}$ \\
        \bottomrule
    \end{tabular}
    \vspace{2mm}
    \caption{Pretraining data size and computational resources used for EXAONE 4.0 language models. EXAONE 4.0 utilizes nearly twice the data of its predecessor, EXAONE 3.5.}
    \label{tab:pretraining}
\end{table}

Furthermore, as recent studies showed that reasoning performance was significantly influenced by the cognitive behavior \citep{gandhi2025cognitivebehaviorsenableselfimproving} acquired from documents seen during pretraining, we perform rigorous data curation during pretraining to enhance post-training performance.

\subsection{Context Length Extension}

In the EXAONE 4.0 model, the maximum context length is extended to 128K tokens. To achieve this, we undertake a two-stage context length extension process. Initially, a model pretrained with a context length of 4K tokens is firstly extended to 32K tokens. Subsequently, it is further extended to 128K tokens.

The long-context fine-tuning process is meticulously executed, with the Needle In A Haystack (NIAH) test~\cite{niah} at each stage to ensure thorough validation of the model's performance. This iterative refinement continues until comprehensive optimization is achieved and the ``green light'' signal is consistently observed across all segments, signifying the successful extension of the context length to 128K tokens without compromising the model's overall performance. 

For the 1.2B model, the context length is extended up to 64K tokens, which is approximately twice as long as the typical maximum length of 32K tokens supported by most models in the 1B-parameter range.

\subsection{Post-training}

\begin{figure}[t!]
    \centering
    \includegraphics[width=\textwidth]{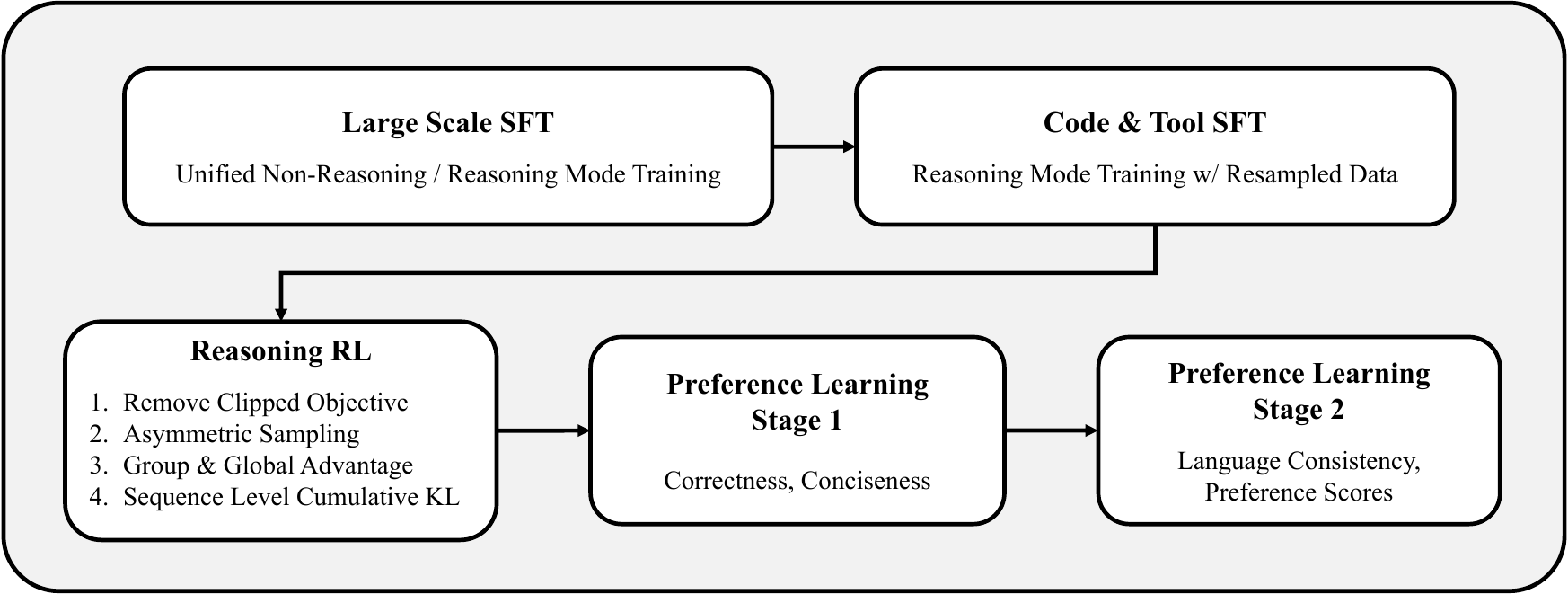}
    \caption{The post-training pipeline of the EXAONE 4.0. The pipeline consists of five stages, which include supervised fine-tuning (SFT), reinforcement learning (RL), and preference learning.}
    \label{fig:posttraining_pipeline}
\end{figure}

In EXAONE 4.0, multiple stages of training is undertaken to enable the model to respond to a variety of user instructions and integrate \textsc{Non-reasoning} and \textsc{Reasoning} models effectively. The training process is primarily organized into three stages: supervised fine-tuning (SFT), reasoning reinforcement learning (RL), and preference learning to integrate \textsc{Non-reasoning} and \textsc{Reasoning} modes as illustrated in Figure~\ref{fig:posttraining_pipeline}.

A significant feature of the post-training phase is the large-scale expansion of the SFT data to enhance performance in an efficient manner. To improve reasoning capabilities, RL is employed. Additionally, a hybrid reward mechanism is used in a two-stage preference learning process to seamlessly integrate \textsc{Non-reasoning} and \textsc{Reasoning} modes.

\subsubsection{Large-scale Supervised Fine-tuning}

The composition of the SFT dataset is divided into non-reasoning and reasoning data. Furthermore, it can be classified into five distinct domains: World Knowledge, Math/Code/Logic, Agentic Tool Use, Long Context, and Multilinguality. Data collection and generation strategies were differentiated for each purpose and domain, and the detailed methodologies are described below.

\paragraph{World Knowledge} For the world knowledge domain, which encompasses a wide range of fields and levels of difficulty, it is essential to enable the distillation of extensive knowledge. Therefore, we filtered problems collected from web sources based on their educational value, prioritizing the use of high-quality data. Among these, we also sample specialized and high-difficulty data to utilize in training for \textsc{Reasoning} mode.

\paragraph{Math, Code, Logic} For the Math, Code, and Logic tasks, the number of unique problems is relatively limited compared to their importance. This is primarily because establishing accurate ground truth is not only essential but also difficult in these domains, thereby limiting our ability to construct as many high-quality problems as desired. Consequently, rather than create unverifiable problems, we train on diverse responses for queries with verifiable answers, and observe that generating multiple responses per unique query is as effective as increasing the diversity or number of unique queries themselves. Furthermore, in the \textsc{Reasoning} mode, responses for Math and Code domains tend to be longer, which increases the risk of degeneration and language inconsistency; thus, careful filtering is applied. Additionally, for the Code domain, we extend our data collection beyond problem-solving to include a software engineering dataset focused on full stack development, created from code corpora.

\paragraph{Long Context} We construct a long-context SFT dataset from web corpora, focusing on tasks that require comprehensive understanding of extended inputs. To train models to identify and reason over dispersed information, we systematically vary both the context length and the location of key content. The dataset also includes instruction-following queries for long-form generation, allowing models to produce coherent and well-structured long outputs. For Korean, we curate long-context data by refining documents such as legal, administrative, and technical texts. These documents are then restructured to accommodate a diverse range of long-context input formats, ensuring variation in structure and content scope.

\paragraph{Agentic Tool Use} To enhance the model’s capability for agentic tool use, we construct datasets focused on both single-turn and multi-turn tasks, leveraging diverse tool lists. Rather than merely creating datasets for single tool calls, we emphasize the construction of more complex, long-horizon tool-calling data. Accordingly, we develop user-agent conversations that incorporate user interaction, execution feedback from the environment, and iterative reasoning, ultimately guiding the agent to achieve the user’s desired goal. These datasets are organized in multi-step and multi-turn formats to better support the learning of agentic tool use.

\paragraph{Multilinguality} To support both Korean and Spanish, we construct datasets that not only target cultural and historical knowledge specific to each language, but also enable the model to engage in fluent, natural conversations with users. We create new instructions in both languages and additionally leveraged translations of selected existing samples as queries. For Korean, in particular, we curate data to address topics relevant to local education and industry experts, ensuring that the model is well-equipped to handle domain-specific queries from Korean users.

\paragraph{Unified Mode Training}
In the combined dataset, the \textsc{Non-Reasoning} data primarily consists of diverse tasks, while the \textsc{Reasoning} data is centered on Math and Code domains. Rather than fine-tuning the two modes sequentially, we combine both modes and train them together. The ratio between the two modes is determined by the amount of Reasoning mode data. If the token ratio of \textsc{Reasoning} mode is too high, we observe that the model tends to behave as if it is in \textsc{Reasoning} mode even when \textsc{Non-Reasoning} mode is enabled. Through ablation studies, we set the token ratio of \textsc{Reasoning} to \textsc{Non-Reasoning} data to 1.5:1.

After unified \textsc{Non-Reasoning/Reasoning} mode fine-tuning, to address domain imbalance, we perform a second round of training using high-quality \textsc{Reasoning} data from the Code and Tool Use domains, reusing these samples to further enhance the performance.

\subsubsection{Reasoning Reinforcement Learning}

To enhance the model's reasoning capabilities, we conduct online reinforcement learning (RL) following supervised fine-tuning (SFT). Previous studies demonstrated that combining the GRPO (Group Relative Policy Optimization) algorithm~\citep{shao2024deepseekmathpushinglimitsmathematical} with verifiable rewards~\citep{lambert2025tulu3pushingfrontiers} can effectively improve model performance. To address the limitations of existing GRPO, we propose a new algorithm, named AGAPO (Asymmetric Sampling and Global Advantage Policy Optimization).

Our training dataset encompasses curated data across four categories: mathematics, code, science, and instruction following. To focus training on more informative data samples, we perform accuracy-based filtering by generating eight responses from the SFT model and excluding samples where all eight responses are correct, a pre-filtering step that removes problems that are easy for the model to avoid inefficient training.

The reward function used in RL is tailored for each category. For the mathematics category, a rule-based verifier is used to determine correctness. In the code category, a response is considered correct if its final code block passes all associated test cases. For the science category, a rule-based verifier is first applied; if a response is deemed incorrect, an LLM-judge then performs a more flexible verification. Finally, for the instruction following category, a reward of 1 is assigned if all constraints are satisfied, and 0 otherwise.

For the algorithm design, AGAPO comprehensively improves upon existing methods. Its main features are as follows:

\begin{itemize}
  \item \textbf{Remove Clipped Objective.}
        Previous research has questioned the necessity of PPO(Proximal Policy Optimization)~\citep{schulman2017proximalpolicyoptimizationalgorithms}’s  clip loss~\citep{ahmadian2024basicsrevisitingreinforcestyle} and shown that this clipped objective can degrade performance~\citep{minimax2025minimaxm1scalingtesttimecompute} by preventing crucial, low-probability tokens from contributing to gradient updates. These tokens are often associated with reflective behaviors that serve as forks in the reasoning path. AGAPO removes the clipping from PPO and instead uses a standard policy gradient loss. This approach is designed to prevent the dropping of these exploratory tokens, allowing for more substantial policy updates while maintaining training stability.
  \item \textbf{Asymmetric Sampling.}
        Previous works filter out samples where all responses were either correct or incorrect~\citep{yu2025dapoopensourcellmreinforcement,he2025skyworkopenreasoner1} because they result in a zero advantage for GRPO. However, as recent work has shown the effectiveness of Negative Sample Reinforcement~\citep{zhu2025surprisingeffectivenessnegativereinforcement}, AGAPO utilizes an asymmetric sampling method that does not discard samples where all responses are incorrect, thereby including a higher proportion of negative feedback. For these all-incorrect samples, a small negative reward is assigned through the advantage calculation, allowing them to be used to guide the model away from erroneous reasoning paths.
  \item \textbf{Group\&Global Advantages.}
        GRPO advantage method does not account for the distribution of the entire batch, which makes it difficult to assign appropriate negative rewards to groups of all-incorrect samples. To improve this, AGAPO calculates the advantage in two stages: group and global. First, at the group level, the advantage is computed using the Leave-One-Out(LOO) method~\citep{ahmadian2024basicsrevisitingreinforcestyle} within a response group based on verifiable reward accuracy. Next, normalization is performed across the entire batch (global) to calculate a final advantage that considers the full batch distribution.
  \item \textbf{Sequence Level Cumulative KL.}
        To enhance performance while preserving capabilities learned during the SFT stage, a KL penalty is applied. We adopt the sequence-level cumulative KL~\citep{tang2025pitfallskldivergencegradient}, as proposed in prior research, to ensure the model receives an appropriate gradient during training.
\end{itemize}

\paragraph{Objective}\label{sec:agapo_objective}
The AGAPO objective is defined for a question \(q\) sampled from the training distribution \(P(Q)\). For each question, the current policy \(\pi_{\theta}(\cdot\mid q)\) generates a \emph{group} of \(G\) candidate responses, denoted as \(O=\{o_1,\dots,o_G\}\). Each response \(o_i\) is assigned a verifiable reward \(r_i\in[0,1]\). The objective function maximizes the following:
\begin{equation}
\mathcal{J}_{\mathrm{AGAPO}}(\theta)=
\mathbb{E}_{q\sim P(Q),\, \{o_i\}_{i=1}^{G}\sim\pi_{\theta}(O\mid q)}
\Bigl[
  \frac{1}{G}\sum_{i=1}^{G}\Bigl(
      A_{\mathrm{global},i}\,\log\pi_{\theta}(o_i\mid q)
      -\beta\,D_{\mathrm{KL}}\!\bigl(\pi_{\theta},\pi_{\mathrm{ref}}\bigr)
  \Bigr)
\Bigr].
\end{equation}

The global advantage \(A_{\mathrm{global},i}\) in the objective is calculated in two stages. First, a leave-one-out (LOO) advantage is computed within each group. This advantage is then normalized across the entire mini-batch of size \(K=B\times G\) to yield the final global advantage:

\begin{equation}
A_{\mathrm{loo},i}=r_i-\frac{1}{G-1}\sum_{j\neq i} r_j,
\quad
A_{\mathrm{global},i}=
\frac{A_{\mathrm{loo},i}-\operatorname{mean}\!\bigl(\{A_{\mathrm{loo},k}\}_{k}\bigr)}
{\operatorname{std}\!\bigl(\{A_{\mathrm{loo},k}\}_{k}\bigr)}.
\end{equation}

\subsubsection{Preference Learning with Hybrid Reward}

In the RL stage, we aim to enhance accuracy through verifiable rewards, and do not use human preference. In addition, since the model is specialized for reasoning tasks, we observe a decline in performance in other types of tasks. To overcome these limitations, we introduce an additional preference learning phase.

Preference learning is conducted by directly learning human preferences from chosen and rejected data pairs, akin to the Direct Policy Optimization (DPO) framework~\citep{rafailov2024directpreferenceoptimizationlanguage}. We employ SimPER~\citep{xiao2025simperminimalistapproachpreference}, among various reference-free preference optimization methods for this learning process. The dataset for preference learning is constructed based on on-policy responses~\citep{meng2024simposimplepreferenceoptimization, lambert2025tulu3pushingfrontiers} generated by the model after completing the RL phase. For each query, we generate 4 to 16 responses per task, and select chosen and rejected responses based on a hybrid reward combining verifiable reward, preference reward, language consistency reward, and conciseness reward, tailored per task.

The training is conducted separately for two stages. In the first stage, we focus on increasing token efficiency by reducing the generation length while maintaining the performance of the reasoning mode. Therefore, for reasoning-related verifiable training data, we combine the verifiable reward with a conciseness reward to select the shortest response among the correct answers as the chosen option. In the second stage, we employ a combination of preference reward and language consistency reward for human alignment. For the \textsc{Reasoning} Mode data, preference labeling is performed only on the final answer after the reasoning process is complete. Furthermore, to ensure stability during the second stage of training, a portion of the data from the first stage is sampled and reused.

\subsection{Data Compliance}
\label{sec:data_compliance}

Developing AI models requires a large amount of data, and the acquisition and utilization of this data can lead to various legal issues, such as copyright infringement, intellectual property infringement, and personal information protection violations. To minimize these risks, LG AI Research conducts AI Compliance reviews throughout the entire process of data collection, AI model training, and information provision. For more detailed information, please refer to the EXAONE 3.0 Technical Report~\citep{research2024exaone3078binstruction} and the LG AI Ethics Principles~\citep{lgethics}.

\section{Evaluation}

\begin{table*}[!t]
\centering
\footnotesize
\setlength{\tabcolsep}{4pt}
\resizebox{1.0\textwidth}{!}{
\begin{tabular}{@{}lccccccc@{}}
\toprule
\multicolumn{1}{l}{} & \multicolumn{4}{c}{\makecell{\textsc{Mid-size}}} & \multicolumn{1}{c}{} & \multicolumn{2}{c}{\makecell{\textsc{Frontier}}} \\
\hhline{~====~==} \\
\multicolumn{1}{l|}{} & \multicolumn{1}{c|}{\makecell{\textbf{EXAONE 4.0 32B} \\ \smaller[1]~\textbf{(\textsc{Reasoning})}}} & \makecell{Phi 4 \\ reasoning-plus} & \makecell{Magistral \\ Small-2506} & \makecell{Qwen 3 32B \\ {\smaller[1]~(\textsc{Reasoning})}} & \multicolumn{1}{c}{} & \makecell{Qwen 3 235B \\ {\smaller[1]~(\textsc{Reasoning})}} & \makecell{DeepSeek R1 \\ -0528} \\
\midrule

\multicolumn{1}{l|}{Type} & \multicolumn{1}{c}{Hybrid} & Reasoning & Reasoning & Hybrid & & Hybrid & Reasoning \\
\multicolumn{1}{l|}{\# Total Params} & \multicolumn{1}{c|}{32.0 B} & 14.7 B & 23.6 B & 32.8 B & & 235 B & 671 B \\
\midrule

\rowcolor[rgb]{0.9,0.9,0.9}\multicolumn{8}{c}{\textit{World Knowledge}} \\
\midrule

\multicolumn{1}{l|}{ \textsc{MMLU-Redux}} & \multicolumn{1}{c|}{92.3} & 90.8~~ & 86.8~~ & 90.9$^*$ & & 92.7$^*$ & 93.4$^*$ \\
\multicolumn{1}{l|}{ \textsc{MMLU-Pro}} & \multicolumn{1}{c|}{81.8} & 76.0$^*$ & 73.4~~ & 80.0~~ & & 83.0~~ & 85.0$^*$ \\
\multicolumn{1}{l|}{ \textsc{GPQA-Diamond}} & \multicolumn{1}{c|}{75.4} & 68.9$^*$ & 68.2$^*$ & 68.4$^*$ & & 71.1$^*$ & 81.0$^*$ \\
\midrule

\rowcolor[rgb]{0.9,0.9,0.9}\multicolumn{8}{c}{\textit{Math / Coding}} \\
\midrule

\multicolumn{1}{l|}{ \textsc{AIME 2025}} & \multicolumn{1}{c|}{85.3} & 78.0$^*$ & 62.8$^*$ & 72.9$^*$ & & 81.5$^*$ & 87.5$^*$ \\
\multicolumn{1}{l|}{ \textsc{HMMT Feb 2025}} & \multicolumn{1}{c|}{72.9} & 53.6$^*$ & 43.5~~ & 50.4~~ & & 62.5$^*$ & 79.4$^*$ \\
\multicolumn{1}{l|}{ \textsc{LiveCodeBench v5}} & \multicolumn{1}{c|}{72.6} & 51.7~~ & 55.8$^*$ & 65.7$^*$ & & 70.7$^*$ & 75.2$^*$ \\
\multicolumn{1}{l|}{ \textsc{LiveCodeBench v6}} & \multicolumn{1}{c|}{66.7} & 47.1~~ & 47.4$^*$ & 60.1~~ & & 58.9$^*$ & 70.3$^*$ \\
\midrule

\rowcolor[rgb]{0.9,0.9,0.9}\multicolumn{8}{c}{\textit{Instruction Following}} \\
\midrule

\multicolumn{1}{l|}{ \textsc{IFEval}} & \multicolumn{1}{c|}{83.7} & 84.9$^*$ & 37.9~~ & 85.0$^*$ & & 83.4$^*$ & 80.8~~ \\
\multicolumn{1}{l|}{ \textsc{Multi-IF~{\smaller[1]~(EN)}}} & \multicolumn{1}{c|}{73.5} & 56.1~~ & 27.4~~ & 73.4~~ & & 73.4~~ & 72.0~~ \\
\midrule

\rowcolor[rgb]{0.9,0.9,0.9}\multicolumn{8}{c}{\textit{Agentic Tool Use}} \\
\midrule

\multicolumn{1}{l|}{ \textsc{BFCL-v3}} & \multicolumn{1}{c|}{63.9} & N/A~~ & 40.4~~ & 70.3$^*$ & & 70.8$^*$ & 64.7$^*$ \\
\multicolumn{1}{l|}{ \textsc{Tau-Bench} (Airline)} & \multicolumn{1}{c|}{51.5} & N/A~~ & 38.5~~ & 34.5~~ & & 37.5~~ & 53.5$^*$ \\
\multicolumn{1}{l|}{ \textsc{Tau-Bench} (Retail)} & \multicolumn{1}{c|}{62.8} & N/A~~ & 10.2~~ & 55.2~~ & & 58.3~~ & 63.9$^*$ \\
\midrule

\rowcolor[rgb]{0.9,0.9,0.9}\multicolumn{8}{c}{\textit{Multilinguality}} \\
\midrule

\multicolumn{1}{l|}{ \textsc{KMMLU-Pro~{\smaller[1]~(KO)}}} & \multicolumn{1}{c|}{67.7} & 55.8~~ & 51.5~~ & 61.4~~ & & 68.1~~ & 71.7~~ \\
\multicolumn{1}{l|}{ \textsc{KMMLU-Redux~{\smaller[1]~(KO)}}} & \multicolumn{1}{c|}{72.7} & 62.7~~ & 54.6~~ & 67.5~~ & & 74.5~~ & 77.0~~ \\
\multicolumn{1}{l|}{ \textsc{KSM~{\smaller[1]~(KO)}}} & \multicolumn{1}{c|}{87.6} & 79.8~~ & 71.9~~ & 82.8~~ & & 86.2~~ & 86.7~~ \\

\multicolumn{1}{l|}{ \textsc{MMMLU~{\smaller[1]~(ES)}}} & \multicolumn{1}{c|}{85.6} & 84.3~~ & 68.9~~ & 82.8$^*$ & & 86.7$^*$ & 88.2~~ \\
\multicolumn{1}{l|}{ \textsc{MATH500~{\smaller[1]~(ES)}}} & \multicolumn{1}{c|}{95.8} & 94.2~~ & 83.5~~ & 94.3~~ & & 95.1~~ & 96.0~~ \\
\bottomrule

\end{tabular}
}
\caption{The main evaluation results of EXAONE 4.0 32B \textsc{Reasoning} mode. Missing entries (\texttt{N/A}, Not Applicable) indicate that the corresponding model does not support the given input length or task. Asterisk ($^*$) indicates that the scores are from each baseline model's official technical report, blog or leaderboard.}
\label{tab:32b_reasoning}
\end{table*}

\subsection{Benchmarks}

We evaluate EXAONE 4.0 on a diverse set of benchmarks spanning 6 categories: World Knowledge, Math/Coding, Instruction Following, Long Context, Agentic Tool Use, and Multilinguality.

\begin{itemize}
    \item \textbf{World Knowledge} We select benchmarks to evaluate the extent of our model's world knowledge, including \textsc{MMLU-Redux}~\citep{gema2025mmlu} and \textsc{MMLU-Pro}~\citep{wang2024mmluprorobustchallengingmultitask}, a refined and extended version of \textsc{MMLU}~\citep{hendrycks2021measuringmassivemultitasklanguage}. Additionally, we utilize \textsc{GPQA-Diamond}~\citep{rein2023gpqagraduatelevelgoogleproofqa} to assess the expert-level knowledge in Biology, Physics, and Chemistry.
    \item \textbf{Math/Coding} Challenging benchmarks in Math and Coding categories are adopted to evaluate the test-time computational capability of EXAONE 4.0. For Math, we utilize two math Olympiad competitions: \textsc{AIME 2025}~\citep{maa2025aime} and \textsc{HMMT Feb 2025}~\citep{balunovic_srimatharena_2025}. For Coding, \textsc{LiveCodeBench v5} and \textsc{v6}~\citep{jain2024livecodebenchholisticcontaminationfree} are chosen.
    \item \textbf{Instruction Following} To evaluate how well our models understand and align with users' instructions, we select \textsc{IFEval}~\citep{zhou2023instructionfollowingevaluationlargelanguage} and \textsc{Multi-IF~}\citep{he2024multiifbenchmarkingllmsmultiturn}, the latter being an extension of \textsc{IFEval} to support multi-turn and multilingual scenarios. We use only the English subset of \textsc{Multi-IF} to focus on assessing the multi-turn instruction-following ability on English.
    \item \textbf{Long Context} To evaluate the model’s ability to understand and solve tasks requiring long-context comprehension, we adopt three representative benchmarks: \textsc{HELMET}~\citep{yen2025helmetevaluatelongcontextlanguage}, \textsc{RULER}~\citep{hsieh2024rulerwhatsrealcontext}, and \textsc{LongBench}~\citep{bai2024longbenchbilingualmultitaskbenchmark}. These benchmarks collectively cover both synthetic tasks and real-world scenarios. To maintain a coherent evaluation focused on core long-context capabilities, we exclude the \textit{LongCite} task from HELMET (see Appendix~\ref{appendix:evaluation_details} for further details).

    \item \textbf{Agentic Tool Use} With the advancement of LLM-based agents, numerous benchmarks have emerged to evaluate their tool-use capabilities, among which we focus on the two most widely adopted: \textsc{BFCL-v3}~\citep{patil2025bfcl} and \textsc{Tau-Bench}~\citep{yao2024taubenchbenchmarktoolagentuserinteraction}. \textsc{BFCL-v3} evaluates various aspects of function-calling abilities. \textsc{Tau-Bench} assesses tool calling performance through simulated conversations between a user LLM. We utilize \textsl{gpt-4.1-2025-04-14} model as the user role.
    \item \textbf{Multilinguality} Beyond English, we evaluate our models on two additional languages: Korean and Spanish. For Korean, we use \textsc{KMMLU-Pro}\footnote{\url{https://huggingface.co/datasets/LGAI-EXAONE/KMMLU-Pro}} for measuring practical applicability on professional knowledge and \textsc{KMMLU-Redux}\footnote{\url{https://huggingface.co/datasets/LGAI-EXAONE/KMMLU-Redux}} for assessing real-world expert knowledge instead of KMMLU~\citep{son-etal-2025-kmmlu} to ensure benchmark reliability. KMMLU have been reported dataset error and contamination issue between pre-training corpora and task dataset. In addition, we employ Korean School Math (\textsc{KSM}) subset of \textsc{HRM8K}~\citep{ko2025understandsolvetranslatebridging} to evaluate a wide range of Korean mathematical knowledge from high-school to Olympiad level. To evaluate the models' ability to handle long-context Korean inputs, we also include an in-house benchmark, \textsc{Ko-LongBench} (Please refer to Appendix~\ref{appendix:ko_longbench} for details).
    For Spanish, we adopt the translated version of existing benchmarks. To be specific, we use \textsc{MMMLU{\smaller[1]~(es)}}~\footnote{\url{https://huggingface.co/datasets/openai/MMMLU}} and~\textsc{MATH500~\citep{lightman2023lets}{\smaller[1]~(es)}}~\footnote{\url{https://huggingface.co/datasets/bezir/MATH-500-multilingual}}. Furthermore, we assess translation ability using \textsc{WMT24++}~\citep{{deutsch2025wmt24expandinglanguagecoverage}}, a widely-used translation benchmark. We consider only \textsc{EN} and \textsc{ES} pair, and utilize LLM-as-a-judge~\footnote{\textsl{gpt-4.1-2025-04-14} is used for the judge model. We follow reference-based direct assessment method used in WMT24++~\citep{deutsch2025wmt24expandinglanguagecoverage}. The exact prompt used for judge is in Appendix~\ref{appendix:wmt24pp}.} to score the translation quality.
    
\end{itemize}

\begin{table*}[!t]
\centering
\footnotesize
\setlength{\tabcolsep}{2pt}
\resizebox{1.0\textwidth}{!}{
\begin{tabular}{@{}lcccccccccc@{}}
\toprule
\multicolumn{1}{l}{} & \multicolumn{5}{c}{\makecell{\textsc{Mid-size}}} & \multicolumn{1}{c}{} & \multicolumn{3}{c}{\makecell{\textsc{Frontier}}} \\
\hhline{~=====~===} \\
\multicolumn{1}{l|}{} & \multicolumn{1}{c|}{\makecell{\textbf{EXAONE 4.0 32B} \\ \smaller[1]~\textbf{(\textsc{Non-}} \\ \smaller[1]~\textbf{\textsc{reasoning})}}} & \makecell{Phi 4} & \makecell{Mistral \\ Small-2506} & \makecell{Gemma 3 \\ 27B} & \makecell{Qwen 3 32B \\ {\smaller[1]~(\textsc{Non-}}\\{\smaller[1]~\textsc{reasoning})}} & \makecell{} & \makecell{Qwen 3 235B \\ {\smaller[1]~(\textsc{Non-}}\\{\smaller[1]~\textsc{reasoning})}} & \makecell{Llama 4 \\ Maverick} & \makecell{DeepSeek V3 \\ -0324} \\
\midrule

\multicolumn{1}{l|}{Type} & \multicolumn{1}{c|}{Hybrid} & \makecell{Non-\\Reasoning} & \makecell{Non-\\Reasoning} & \makecell{Non-\\Reasoning} & Hybrid & & Hybrid & \makecell{Non-\\Reasoning} & \makecell{Non-\\Reasoning} \\
\multicolumn{1}{l|}{\# Total Params} & \multicolumn{1}{c|}{32.0 B} & 14.7 B & 24.0 B & 27.4B & 32.8 B & & 235 B & 402 B & 671 B \\

\midrule
\rowcolor[rgb]{0.9,0.9,0.9}\multicolumn{10}{c}{\textit{World Knowledge}} \\
\midrule

\multicolumn{1}{l|}{ \textsc{MMLU-Redux}} & \multicolumn{1}{c|}{89.8} & 88.3~~ & 85.9~~ & 85.0~~ & 85.7$^*$ & & 89.2$^*$ & 92.3~~ & 92.3~~ \\
\multicolumn{1}{l|}{ \textsc{MMLU-Pro}} & \multicolumn{1}{c|}{77.6} & 70.4$^*$ & 69.1$^*$ & 67.5$^*$ & 74.4~~ & & 77.4~~ & 80.5$^*$ & 81.2$^*$ \\
\multicolumn{1}{l|}{ \textsc{GPQA-Diamond}} & \multicolumn{1}{c|}{63.7} & 56.1$^*$ & 46.1$^*$ & 42.4$^*$ & 54.6$^*$ & & 62.9$^*$ & 69.8$^*$ & 68.4$^*$ \\
\midrule

\rowcolor[rgb]{0.9,0.9,0.9}\multicolumn{10}{c}{\textit{Math / Coding}} \\
\midrule

\multicolumn{1}{l|}{ \textsc{AIME 2025}} & \multicolumn{1}{c|}{35.9} & 17.8~~ & 30.2~~ & 23.8~~ & 20.2$^*$ & & 24.7$^*$ & 18.0~~ & 50.0$^*$ \\
\multicolumn{1}{l|}{ \textsc{HMMT Feb 2025}} & \multicolumn{1}{c|}{21.8} & 4.0~~ & 16.9~~ & 10.3~~ & 9.8 & & 11.9~~ & 7.3 & 29.2$^*$ \\
\multicolumn{1}{l|}{ \textsc{LiveCodeBench v5}} & \multicolumn{1}{c|}{43.3} & 24.6~~ & 25.8~~ & 27.5~~ & 31.3$^*$ & & 35.3$^*$ & 43.4$^*$ & 46.7~~ \\
\multicolumn{1}{l|}{ \textsc{LiveCodeBench v6}} & \multicolumn{1}{c|}{43.1} & 27.4~~ & 26.9~~ & 29.7~~ & 28.0~~ & & 31.4~~ & 32.7~~ & 44.0~~ \\
\midrule

\rowcolor[rgb]{0.9,0.9,0.9}\multicolumn{10}{c}{\textit{Instruction Following}} \\
\midrule

\multicolumn{1}{l|}{ \textsc{IFEval}} & \multicolumn{1}{c|}{84.8} & 63.0$^*$ & 77.8~~ & 82.6~~ & 83.2$^*$ & & 83.2$^*$ & 85.4~~ & 81.2~~ \\
\multicolumn{1}{l|}{ \textsc{Multi-IF~{\smaller[1]~(EN)}}} & \multicolumn{1}{c|}{71.6} & 47.7~~ & 63.2~~ & 72.1~~ & 71.9~~ & & 72.5~~ & 77.9~~ & 68.3~~ \\
\midrule

\rowcolor[rgb]{0.9,0.9,0.9}\multicolumn{10}{c}{\textit{Long Context}} \\
\midrule

\multicolumn{1}{l|}{ \textsc{HELMET}} & \multicolumn{1}{c|}{58.3} & N/A~~ & 61.9~~ & 58.3$^*$ & 54.5~~ & & 63.3~~ & 13.7~~ & N/A~~ \\
\multicolumn{1}{l|}{ \textsc{RULER}} & \multicolumn{1}{c|}{88.2} & N/A~~ & 71.8~~ & 66.0$^*$ & 85.6$^*$ & & 90.6$^*$ & 2.9 & N/A~~ \\
\multicolumn{1}{l|}{ \textsc{LongBench v1}} & \multicolumn{1}{c|}{48.1} & N/A~~ & 51.5~~ & 51.5~~ & 44.2~~ & & 45.3~~ & 34.7~~ & N/A~~ \\
\midrule

\rowcolor[rgb]{0.9,0.9,0.9}\multicolumn{10}{c}{\textit{Agentic Tool Use}} \\
\midrule

\multicolumn{1}{l|}{ \textsc{BFCL-v3}} & \multicolumn{1}{c|}{65.2} & N/A~~ & 57.7$^*$ & N/A~~ & 63.0$^*$ & & 68.0$^*$ & 52.9$^*$ & 63.8$^*$ \\
\multicolumn{1}{l|}{ \textsc{Tau-Bench} (Airline)} & \multicolumn{1}{c|}{25.5} & N/A~~ & 36.1~~ & N/A~~ & 16.0~~ & & 27.0~~ & 38.0~~ & 40.5~~ \\
\multicolumn{1}{l|}{ \textsc{Tau-Bench} (Retail)} & \multicolumn{1}{c|}{55.9} & N/A~~ & 35.5~~ & N/A~~ & 47.6~~ & & 56.5~~ & 6.5 & 68.5~~ \\
\midrule

\rowcolor[rgb]{0.9,0.9,0.9}\multicolumn{10}{c}{\textit{Multilinguality}} \\
\midrule

\multicolumn{1}{l|}{ \textsc{KMMLU-Pro~{\smaller[1]~(KO)}}} & \multicolumn{1}{c|}{60.0} & 44.8~~ & 51.0~~ & 50.7~~ & 58.3~~ & & 64.4~~ & 68.8~~ & 67.3~~ \\
\multicolumn{1}{l|}{ \textsc{KMMLU-Redux~{\smaller[1]~(KO)}}} & \multicolumn{1}{c|}{64.8} & 50.1~~ & 53.6~~ & 53.3~~ & 64.4~~ & & 71.7~~ & 76.9~~ & 72.2~~ \\
\multicolumn{1}{l|}{ \textsc{KSM~{\smaller[1]~(KO)}}} & \multicolumn{1}{c|}{59.8} & 29.1~~ & 35.5~~ & 36.1~~ & 41.3~~ & & 46.6~~ & 40.6~~ & 63.5~~ \\
\multicolumn{1}{l|}{ \textsc{Ko-LongBench~{\smaller[1]~(KO)}}} & \multicolumn{1}{c|}{76.9} & N/A~~ & 55.4~~ & 72.0~~ & 73.9~~ & & 74.6~~ & 65.6~~ & N/A~~ \\
\multicolumn{1}{l|}{ \textsc{MMMLU~{\smaller[1]~(ES)}}} & \multicolumn{1}{c|}{80.6} & 81.2~~ & 78.4~~ & 78.7~~ & 82.1$^*$ & & 83.7$^*$ & 86.9~~ & 86.7~~ \\
\multicolumn{1}{l|}{ \textsc{MATH500~{\smaller[1]~(ES)}}} & \multicolumn{1}{c|}{87.3} & 78.2~~ & 83.4~~ & 86.8~~ & 84.7~~ & & 87.2~~ & 78.7~~ & 89.2~~ \\
\multicolumn{1}{l|}{ \textsc{WMT24++~{\smaller[1]~(ES)}}} & \multicolumn{1}{c|}{90.7} & 89.3~~ & 92.2~~ & 93.1~~ & 91.4~~ & & 92.9~~ & 92.7~~ & 94.3~~ \\
\bottomrule

\end{tabular}
}
\caption{The main evaluation results of EXAONE 4.0 32B \textsc{Non-reasoning} mode. Missing entries (\texttt{N/A}, Not Applicable) indicate that the corresponding model does not support the given input length or task. Asterisk ($^*$) indicates that the scores are from each baseline model's official technical report, blog or leaderboard.}
\label{tab:32b_non_reasoning}
\end{table*}

\subsection{Baselines}

To evaluate the performance of language models from various perspectives, recently released open-weight models are selected as baseline models. These baseline models include not only models of similar sizes but also frontier-level models exceeding 100B parameters, which exhibits superior performance. These models can be divided into three types: (1) Non-Reasoning models, which generate their responses in CoT (Chain-of-Thought) style, (2) Reasoning models, which generate in long CoT style, and (3) Hybrid model, which generate in either CoT or long CoT style depending on the mode. Detailed information about the models is presented in the Appendix~\ref{appendix:baseline_model}.

\subsection{Experimental Setup}

\paragraph{Hyperparameters} We sample $n$ different responses for each problem in benchmarks with limited examples to ensure evaluation stability. Specifically, we sample $n=8$ responses for \textsc{GPQA-Diamond}, $n=32$ for \textsc{AIME 2025} and \textsc{HMMT Feb 2025}, and $n=4$ for \textsc{LiveCodeBench v5/6}, \textsc{Tau-Bench} and \textsc{MATH500{\smaller[1]~(es)}}. The accuracy is averaged over the $n$ samples.
In \textsc{Reasoning} mode, we set temperature to 0.6, top-p~\citep{Holtzman2020The} to 0.95, and apply a presence penalty of 1.5 only for our 32B model. In contrast, for \textsc{Non-Reasoning} mode, greedy decoding is used for a single ($n=1$) generated response, while the same sampling settings as \textsc{Reasoning} mode (except with a presence penalty of 0.0) are used when generating $n>1$ responses.
We generate a maximum of 64K tokens for \textsc{AIME 2025}, \textsc{HMMT Feb 2025}, \textsc{LiveCodeBench v5/6}, and \textsc{KSM} benchmarks, while 32K for other benchmarks.

\paragraph{Long-Context Evaluation of \textsc{Small-Size} models} In evaluating long-context performance of \textsc{Small-Size} \textsc{NON-REASONING} models, we extend the context lengths of Qwen3 1.7B and Qwen3 0.6B beyond their 32K token limit by applying YaRN~\citep{peng2024yarn}, enabling inference up to 64K tokens. For reference, evaluation results of models such as Gemma-3-1B, EXAONE-3.5-2.4B-Instruct, Qwen3 1.7B, and Qwen3 0.6B at context lengths up to 32K tokens are provided in the Appendix~\ref{appendix:evaluation_details}.

\paragraph{Baselines Reproduction} For baseline models, we borrow scores reported in each model's official technical report, blog, or leaderboard\footnote{We refer to \url{https://github.com/LiveCodeBench/submissions} for \textsc{LiveCodeBench}, \url{https://matharena.ai/} for \textsc{HMMT Feb 2025}, and \url{https://gorilla.cs.berkeley.edu/leaderboard.html} for \textsc{BFCL-v3}.} if available. If not, we reproduce the results in our evaluation environment, following the recommended settings when they are explicitly stated\footnote{For example, the Qwen3 series explicitly specifies recommended decoding parameters in its Hugging Face repository.}.

\subsection{Experimental Results}

\begin{table*}[!t]
\centering
\footnotesize
\resizebox{0.95\textwidth}{!}{
\begin{tabular}{@{}lcccccc@{}}
\toprule
\multicolumn{1}{l}{} & \multicolumn{5}{c}{\makecell{\textsc{Small-size}}} \\
\hhline{~=====} \\
\multicolumn{1}{l|}{} & \multicolumn{1}{c|}{\makecell{\textbf{EXAONE 4.0 1.2B} \\ \smaller[1]~\textbf{(\textsc{Reasoning})}}} & \makecell{EXAONE Deep\\2.4B} & \makecell{Qwen 3 0.6B \\ \smaller[1](\textsc{Reasoning})} & \makecell{Qwen 3 1.7B \\ \smaller[1]~(\textsc{Reasoning})} & \makecell{SmolLM 3 3B \\ \smaller[1]~(\textsc{Reasoning})} \\
\midrule

\multicolumn{1}{l|}{Type} & \multicolumn{1}{c|}{Hybrid} & Reasoning & Hybrid &  Hybrid & Hybrid \\
\multicolumn{1}{l|}{\# Total Params} & \multicolumn{1}{c|}{1.28 B} & 2.41 B & 596 M & 1.72 B & 3.08 B \\
\midrule

\rowcolor[rgb]{0.9,0.9,0.9}\multicolumn{6}{c}{\textit{World Knowledge}} \\
\midrule

\multicolumn{1}{l|}{ \textsc{MMLU-Redux}} & \multicolumn{1}{c|}{71.5} & 68.9~~ & 55.6$^*$ & 73.9$^*$ & 74.8~~ \\
\multicolumn{1}{l|}{ \textsc{MMLU-Pro}} & \multicolumn{1}{c|}{59.3} & 56.4$^*$ & 38.3~~ & 57.7~~ & 57.8~~ \\
\multicolumn{1}{l|}{ \textsc{GPQA-Diamond}} & \multicolumn{1}{c|}{52.0} & 54.3$^*$ & 27.9$^*$ & 40.1$^*$ & 41.7$^*$ \\
\midrule

\rowcolor[rgb]{0.9,0.9,0.9}\multicolumn{6}{c}{\textit{Math / Coding}} \\
\midrule

\multicolumn{1}{l|}{ \textsc{AIME 2025}} & \multicolumn{1}{c|}{45.2} & 47.9$^*$ & 15.1$^*$ & 36.8$^*$ & 36.7$^*$ \\
\multicolumn{1}{l|}{ \textsc{HMMT Feb 2025}} & \multicolumn{1}{c|}{34.0} & 27.3~~ & 7.0 & 21.8~~ & 26.0~~ \\
\multicolumn{1}{l|}{ \textsc{LiveCodeBench v5}} & \multicolumn{1}{c|}{44.6} & 47.2~~ & 12.3$^*$ & 33.2$^*$ & 27.6~~ \\
\multicolumn{1}{l|}{ \textsc{LiveCodeBench v6}} & \multicolumn{1}{c|}{45.3} & 43.1~~ & 16.4~~ & 29.9~~ & 29.1~~ \\
\midrule

\rowcolor[rgb]{0.9,0.9,0.9}\multicolumn{6}{c}{\textit{Instruction Following}} \\
\midrule

\multicolumn{1}{l|}{ \textsc{IFEval}} & \multicolumn{1}{c|}{67.8} & 71.0~~ & 59.2$^*$ & 72.5$^*$ & 71.2$^*$ \\
\multicolumn{1}{l|}{ \textsc{Multi-IF~{\smaller[1]~(EN)}}} & \multicolumn{1}{c|}{53.9} & 54.5~~ & 37.5~~ & 53.5~~ & 47.5~~ \\
\midrule

\rowcolor[rgb]{0.9,0.9,0.9}\multicolumn{6}{c}{\textit{Agentic Tool Use}} \\
\midrule

\multicolumn{1}{l|}{ \textsc{BFCL-v3}} & \multicolumn{1}{c|}{52.9} & N/A~~ & 46.4$^*$ & 56.6$^*$ & 37.1~~ \\
\multicolumn{1}{l|}{ \textsc{Tau-Bench} (Airline)} & \multicolumn{1}{c|}{20.5} & N/A~~ & 22.0~~ & 31.0~~ & 37.0~~ \\
\multicolumn{1}{l|}{ \textsc{Tau-Bench} (Retail)} & \multicolumn{1}{c|}{28.1} & N/A~~ & 3.3 & 6.5 & 5.4 \\
\midrule

\rowcolor[rgb]{0.9,0.9,0.9}\multicolumn{6}{c}{\textit{Multilinguality}} \\
\midrule

\multicolumn{1}{l|}{ \textsc{KMMLU-Pro~{\smaller[1]~(KO)}}} & \multicolumn{1}{c|}{42.7} & 24.6~~ & 21.6~~ & 38.3~~ & 30.5~~ \\
\multicolumn{1}{l|}{ \textsc{KMMLU-Redux~{\smaller[1]~(KO)}}} & \multicolumn{1}{c|}{46.9} & 25.0~~ & 24.5~~ & 38.0~~ & 33.7~~ \\
\multicolumn{1}{l|}{ \textsc{KSM~{\smaller[1]~(KO)}}} & \multicolumn{1}{c|}{60.6} & 60.9~~ & 22.8~~ & 52.9~~ & 49.7~~ \\

\multicolumn{1}{l|}{ \textsc{MMMLU~{\smaller[1]~(ES)}}} & \multicolumn{1}{c|}{62.4} & 51.4~~ & 48.8$^*$ & 64.5$^*$ & 64.7~~ \\
\multicolumn{1}{l|}{ \textsc{MATH500~{\smaller[1]~(ES)}}} & \multicolumn{1}{c|}{88.8} & 84.5~~ & 70.6~~ & 87.9~~ & 87.5~~ \\

\bottomrule

\end{tabular}
}
\caption{The main evaluation results of EXAONE 4.0 1.2B \textsc{Reasoning} mode. Missing entries (\texttt{N/A}, Not Applicable) indicate that the corresponding model does not support the given input length or task. Asterisk ($^*$) indicates that the scores are from each baseline model's official technical report, blog or leaderboard.}
\label{tab:1b_reasoning}
\end{table*}
\begin{table*}[!t]
\centering
\footnotesize
\resizebox{1.0\textwidth}{!}{
\begin{tabular}{@{}lccccc@{}}
\toprule
\multicolumn{1}{l}{} & \multicolumn{5}{c}{\makecell{\textsc{Small-size}}} \\
\hhline{~=====} \\
\multicolumn{1}{l|}{} & \multicolumn{1}{c|}{\makecell{\textbf{EXAONE 4.0 1.2B} \\ \smaller[1]~\textbf{(\textsc{Non-reasoning})}}} & \makecell{Qwen 3 0.6B \\ \smaller[1](\textsc{Non-reasoning})} & \makecell{Gemma 3 1B} & \makecell{Qwen 3 1.7B \\ \smaller[1]~(\textsc{Non-reasoning})} & \makecell{SmolLM 3 3B \\ \smaller[1]~(\textsc{Non-reasoning})} \\
\midrule

\multicolumn{1}{l|}{Type} & \multicolumn{1}{c|}{Hybrid} & Hybrid & Non-Reasoning &  Hybrid & Hybrid \\
\multicolumn{1}{l|}{\# Total Params} & \multicolumn{1}{c|}{1.28 B} & 596 M & 1.00 B & 1.72 B & 3.08 B \\
\midrule

\rowcolor[rgb]{0.9,0.9,0.9}\multicolumn{6}{c}{\textit{World Knowledge}} \\
\midrule

\multicolumn{1}{l|}{ \textsc{MMLU-Redux}} & \multicolumn{1}{c|}{66.9} & 44.6$^*$ & 40.9~~ & 63.4$^*$ & 65.0~~ \\
\multicolumn{1}{l|}{ \textsc{MMLU-Pro}} & \multicolumn{1}{c|}{52.0} & 26.6~~ & 14.7$^*$ & 43.7~~ & 43.6~~ \\
\multicolumn{1}{l|}{ \textsc{GPQA-Diamond}} & \multicolumn{1}{c|}{40.1} & 22.9$^*$ & 19.2$^*$ & 28.6$^*$ & 35.7$^*$ \\
\midrule

\rowcolor[rgb]{0.9,0.9,0.9}\multicolumn{6}{c}{\textit{Math / Coding}} \\
\midrule

\multicolumn{1}{l|}{ \textsc{AIME 2025}} & \multicolumn{1}{c|}{23.5} & ~~2.6$^*$ & 2.1 & ~~9.8$^*$ & ~~9.3$^*$ \\
\multicolumn{1}{l|}{ \textsc{HMMT Feb 2025}} & \multicolumn{1}{c|}{13.0} & ~~1.0~~ & 1.5 & 5.1 & 4.7 \\
\multicolumn{1}{l|}{ \textsc{LiveCodeBench v5}} & \multicolumn{1}{c|}{26.4} & ~~3.6$^*$ & 1.8 & 11.6$^*$ & 11.4~~ \\
\multicolumn{1}{l|}{ \textsc{LiveCodeBench v6}} & \multicolumn{1}{c|}{30.1} & ~~6.9~~ & 2.3 & 16.6~~ & 20.6~~ \\
\midrule

\rowcolor[rgb]{0.9,0.9,0.9}\multicolumn{6}{c}{\textit{Instruction Following}} \\
\midrule

\multicolumn{1}{l|}{ \textsc{IFEval}} & \multicolumn{1}{c|}{74.7} & 54.5$^*$ & 80.2$^*$ & 68.2$^*$ & 76.7$^*$ \\
\multicolumn{1}{l|}{ \textsc{Multi-IF~{\smaller[1]~(en)}}} & \multicolumn{1}{c|}{62.1} & 37.5~~ & 32.5~~ & 51.0~~ & 51.9~~ \\
\midrule

\rowcolor[rgb]{0.9,0.9,0.9}\multicolumn{6}{c}{\textit{Long Context}} \\
\midrule

\multicolumn{1}{l|}{ \textsc{HELMET}} & \multicolumn{1}{c|}{41.2} & 21.1~~ & N/A~~ & 33.8~~ & 38.6~~ \\
\multicolumn{1}{l|}{ \textsc{RULER}} & \multicolumn{1}{c|}{77.4} & 55.1~~ & N/A~~ & 65.9~~ & 66.3~~ \\
\multicolumn{1}{l|}{ \textsc{LongBench v1}} & \multicolumn{1}{c|}{36.9} & 32.4~~ & N/A~~ & 41.9~~ & 39.9~~ \\
\midrule

\rowcolor[rgb]{0.9,0.9,0.9}\multicolumn{6}{c}{\textit{Agentic Tool Use}} \\
\midrule

\multicolumn{1}{l|}{ \textsc{BFCL-v3}} & \multicolumn{1}{c|}{55.7} & 44.1$^*$ & N/A~~ & 52.2$^*$ & 47.3~~ \\
\multicolumn{1}{l|}{ \textsc{Tau-Bench} (Airline)} & \multicolumn{1}{c|}{10.0} & 31.5~~ & N/A~~ & 13.5~~ & 38.0~~ \\
\multicolumn{1}{l|}{ \textsc{Tau-Bench} (Retail)} & \multicolumn{1}{c|}{21.7} & 5.7 & N/A~~ & 4.6 & 6.7 \\
\midrule

\rowcolor[rgb]{0.9,0.9,0.9}\multicolumn{6}{c}{\textit{Multilinguality}} \\
\midrule

\multicolumn{1}{l|}{ \textsc{KMMLU-Pro~{\smaller[1]~(KO)}}} & \multicolumn{1}{c|}{37.5} & 24.6~~ & 9.7 & 29.5~~ & 27.6~~ \\
\multicolumn{1}{l|}{ \textsc{KMMLU-Redux~{\smaller[1]~(KO)}}} & \multicolumn{1}{c|}{40.4} & 22.8~~ & 19.4~~ & 29.8~~ & 26.4~~ \\
\multicolumn{1}{l|}{ \textsc{KSM~{\smaller[1]~(KO)}}} & \multicolumn{1}{c|}{26.3} & 0.1 & 22.8~~ & 16.3~~ & 16.1~~ \\
\multicolumn{1}{l|}{ \textsc{Ko-LongBench~{\smaller[1]~(KO)}}} & \multicolumn{1}{c|}{69.8} & 16.4~~ & N/A~~ & 57.1~~ & 15.7~~ \\
\multicolumn{1}{l|}{ \textsc{MMMLU~{\smaller[1]~(ES)}}} & \multicolumn{1}{c|}{54.6} & 39.5$^*$ & 35.9~~ & 54.3$^*$ & 55.1~~ \\
\multicolumn{1}{l|}{ \textsc{MATH500~{\smaller[1]~(ES)}}} & \multicolumn{1}{c|}{71.2} & 38.5~~ & 41.2~~ & 66.0~~ & 62.4~~ \\
\multicolumn{1}{l|}{ \textsc{WMT24++~{\smaller[1]~(ES)}}} & \multicolumn{1}{c|}{65.9} & 58.2~~ & 76.9~~ & 76.7~~ & 84.0~~ \\
\bottomrule

\end{tabular}
}
\caption{The main evaluation results of EXAONE 4.0 1.2B \textsc{Non-reasoning} mode. Missing entries (\texttt{N/A}, Not Applicable) indicate that the corresponding model does not support the given input length or task. Asterisk ($^*$) indicates that the scores are from each baseline model's official technical report, blog, or leaderboard.}
\label{tab:1b_non_reasoning}
\end{table*}

Table~\ref{tab:32b_reasoning},~\ref{tab:32b_non_reasoning},~\ref{tab:1b_reasoning}, and~\ref{tab:1b_non_reasoning} present the benchmark performances of our EXAONE 4.0 models in both \textsc{Reasoning} and \textsc{Non-Reasoning} modes.
The key results are summarized below:

\paragraph{Superiority in Math/Coding domains} EXAONE 4.0 models demonstrate extraordinary performance in Math/Coding benchmarks. Specifically, EXAONE 4.0 32B model outperforms Qwen3 235B in both \textsc{Reasoning} and \textsc{Non-Reasoning} modes across all Math/Coding benchmarks. At the same time, EXAONE 4.0 1.2B model surpasses all baselines, except for EXAONE Deep 2.4B in \textsc{Reasoning} mode.

\paragraph{Competitive Performance in Tool Use Scenarios} EXAONE 4.0 32B model shows competitive performance in tool use compared to baseline models. For example, in \textsc{Reasoning} mode, it demonstrates similar performance to R1-0528 in \textsc{Tau-Bench}, and achieves comparable \textsc{BFCL-v3} results with Qwen 3 235B in \textsc{Non-Reasoning} mode. This is noteworthy considering both baselines are much larger than ours. EXAONE 4.0 1.2B model, despite its small size, achieves the highest performance on \textsc{Tau-Bench} (Retail) compared to the baselines.

\paragraph{World Knowledge and \textsc{GPQA}} Both our models excel in benchmarks in the  World Knowledge category. Despite their relatively smaller size compared to the baselines, they achieve competitive performance. Among the benchmarks, the EXAONE 4.0 models especially demonstrate better performance in \textsc{GPQA-Diamond}. Both EXAONE 4.0 32B and 1.2B models achieve second-highest performance in \textsc{GPQA-Diamond} when \textsc{Reasoning} mode is available.

\subsection{Reasoning Budget}
We control the number of reasoning tokens and observe how performance varies according to the \textit{reasoning budget}. Specifically, while we set the maximum number of tokens to 64K for benchmarks in Math/Coding categories int the main experiments, in this section we vary the number of tokens used for reasoning from 1K to 64K in this section. Similar to~\citep{yang2025qwen3technicalreport}, when the model's generation reaches the maximum token budget, we stop the generation, append the text \textit{"Considering the limited time by the user, I have to give the solution based on the thinking directly now.\textbackslash n</think>\textbackslash n\textbackslash n"}, and proceed to generate the answer part. We use same number of sampled responses per each query as in the main experiments ($n=32$ for \textsc{AIME 2025} and $n=4$ for \textsc{LiveCodeBench v6}) and average the result over $n$ responses. We fix the length of the answer part to 8K.

\begin{table*}[!t]
\centering
\footnotesize
\resizebox{0.6\textwidth}{!}{%
\begin{tabular}{@{}llllllll@{}}
\toprule
Reasoning Budget & 64K & 32K & 16K & 8K & 4K & 2K & 1K \\ \midrule
\rowcolor[rgb]{0.9,0.9,0.9}\multicolumn{8}{c}{EXAONE 4.0 32B} \\ \midrule
\textsc{AIME 2025} & 85.3 & 74.8 & 44.2 & 36.8 & 35.5 & 35.7 & 35.6 \\
\textsc{LiveCodeBench v6} & 66.7 & 67.3 & 53.0 & 47.6 & 46.0 & 45.7 & 44.0 \\ \midrule
\rowcolor[rgb]{0.9,0.9,0.9}\multicolumn{8}{c}{EXAONE 4.0 1.2B} \\ \midrule
\textsc{AIME 2025} & 45.2 & 45.3 & 37.1 & 24.6 & 23.2 & 22.7 & 22.3 \\
\textsc{LiveCodeBench v6} & 45.3 & 43.0 & 40.1 & 38.3 & 34.0 & 33.4 & 29.3 \\ \bottomrule

\end{tabular}%
}
\caption{The results of controlling the \textit{reasoning budget} of EXAONE 4.0 models on \textsc{AIME 2025} and \textsc{LiveCodeBench v6}. The reasoning budget indicates the number of tokens used for reasoning part of the model response. We fix the length of the answer part to 8K.}
\label{tab:reasoning_budget}
\end{table*}

The result is presented in Table~\ref{tab:reasoning_budget}. While a reduced reasoning budget leads to some performance degradation, our EXAONE 4.0 models still demonstrate competitive performance even with a 32K reasoning budgets. Specifically, except for the 32B model on \textsc{AIME 2025}, which shows a 12.3\% decrease in performance, the decrease for others is similar or less than 5\%, maintaining competitive results compared to baseline models.

\section{Limitations} \label{Limitations}

\model{}language models, like all existing language models, have certain limitations and may occasionally generate inappropriate responses. 
The language model generates responses based on the output probability of tokens, and it is determined during learning from training data. While we make every effort to exclude personal, harmful, and biased information from the training data, some problematic content may still be included, potentially leading to undesirable responses. Please note that the text generated by \model{}language models does not reflect the views of LG AI Research.

\begin{itemize}
    \item Inappropriate answers may be generated, which contain personal, harmful or other inappropriate information.
    \item Biased responses may be generated, which are associated with age, gender, race, and so on.
    \item The generated responses rely heavily on statistics from the training data, which can result in the generation of semantically or syntactically incorrect sentences.
    \item Since the models do not reflect the latest information, the responses may be false or contradictory.
\end{itemize}
	
LG AI Research strives to reduce potential risks that may arise from \model{}language models. Users are not allowed to engage in any malicious activities (e.g., keying in illegal information) that may induce the creation of inappropriate outputs violating LG AI's ethical principles when using \model{}language models.

\section{Deployment}

Section~\ref{appendix:license} in the Appendix provides license information for using the \model models. Understanding the license information is essential for the legal utilization of the language model.

\section{Conclusion}

In this technical report, we introduce EXAONE 4.0, which integrates \textsc{non-reasoning} mode and \textsc{reasoning} mode. The key features of EXAONE 4.0 include enhancing the practical usability and reasoning capabilities previously supported in EXAONE 3.5 and EXAONE Deep, consolidating them into a single model, and introducing new functionalities such as agentic tool use and support for Spanish. In terms of performance, EXAONE 4.0 demonstrates superior results compared to models of similar scale and achieves competitive performance even compared to frontier models. As part of our future work, we aim to continuously strengthen usability by gradually expanding the supported languages.

Since the release of EXAONE 3.0, LG AI Research has contributed to the expansion of the research ecosystem by publicly disclosing the model in an open-weight format, and has been continuously improving the model based on user feedback. For any improvement suggestions or business-related inquiries regarding the model, please contact us at \href{mailto:contact_us@lgresearch.ai}{contact\_us@lgresearch.ai}.

\newpage

\appendix

\clearpage
\section{Contributors}
\label{appendix:contributors}
All authors are listed in alphabetical order by last name.

\paragraph{Core Contributors}
Eunbi~Choi, Kibong~Choi, Seokhee~Hong, Junwon~Hwang, Hyojin~Jeon, Hyunjik~Jo, Jiyeon~Jung, Joonkee~Kim, Seonghwan~Kim, Soyeon~Kim, Sunkyoung~Kim, Yireun~Kim, Yongil~Kim, Haeju~Lee, Jinsik~Lee, Kyungmin~Lee, Sangha~Park, Heuiyeen~Yeen, Hyeongu~Yun

\paragraph{Contributors}
Kyunghoon~Bae, Stanley~Jungkyu~Choi, Yemuk~Choi, Kyubeen~Han, Taewan~Hwang, Joonwon~Jang, Kijeong~Jeon, Gerrard~Jeongwon~Jo, Euisoon~Kim, Hyosang~Kim, Jihoon~Kim, Youchul~Kim, Edward~Hwayoung~Lee, Gwangho~Lee, Honglak~Lee, Yongmin~Park, Young~Min~Paik, Youngyong~Park, Sanghyun~Seo, Sihoon~Yang, Sihyuk~Yi

\newpage

\section{Model License}
\label{appendix:license}

\textbf{EXAONE AI Model License Agreement 1.2 - NC} \\
\\
This License Agreement (“Agreement”) is entered into between you (“Licensee”) and LG Management Development Institute Co., Ltd. (“Licensor”), governing the use of the EXAONE AI Model (“Model”). By downloading, installing, copying, or using the Model, you agree to comply with and be bound by the terms of this Agreement. If you do not agree to all the terms, you must not download, install, copy, or use the Model. This Agreement constitutes a binding legal agreement between the Licensee and Licensor. \\
\\
\\
\textbf{1. Definitions} \\
\\
\textbf{1.1 Model:} The artificial intelligence model provided by Licensor, which includes any software, algorithms, machine learning models, or related components supplied by Licensor. This definition extends to encompass all updates, enhancements, improvements, bug fixes, patches, or other modifications that may be provided by Licensor from time to time, whether automatically or manually implemented. \\
\\
\textbf{1.2 Derivatives:} Any modifications, alterations, enhancements, improvements, adaptations, or derivative works of the Model created by Licensee or any third party. This includes changes made to the Model's architecture, parameters, data processing methods, or any other aspect of the Model that results in a modification of its functionality or output. \\ 
\\
\textbf{1.3 Output:} Any data, results, content, predictions, analyses, insights, or other materials generated by the Model or Derivatives, regardless of whether they are in their original form or have been further processed or modified by the Licensee. This includes, but is not limited to, textual or numerical produced directly or indirectly through the use of the Model. \\
\\
\textbf{1.4 Licensor:} LG Management Development Institute Co., Ltd., the owner, developer, and provider of the EXAONE AI Model. The Licensor holds all rights, title, and interest in the Model and is responsible for granting licenses to use the Model under the terms specified in this Agreement. \\
\\
\textbf{1.5 Licensee:} The individual, organization, corporation, academic institution, government agency, or other entity using or intending to use the Model under the terms and conditions of this Agreement. The Licensee is responsible for ensuring compliance with the Agreement by all authorized users who access or utilize the Model on behalf of the Licensee. \\
\\
\\
\textbf{2. License Grant} \\ 
\\
\textbf{2.1 Grant of License:} Subject to the terms and conditions outlined in this Agreement, the Licensor hereby grants the Licensee a limited, non-exclusive, non-transferable, worldwide, and revocable license to: \\
\\
a. Access, download, install, and use the Model solely for research and educational purposes. This includes evaluation, testing, academic research, experimentation, learning, teaching, training and participation in competitions, provided that such participation is in a non-commercial context. Notwithstanding Section 3.1, the Licensee may only provide the Model or Derivatives for a competition if no commercial license is granted to the competition organizer or any third party. \\
\\
b. Publicly disclose research results and findings derived from the use of the Model or Derivatives, including publishing papers or presentations. \\
\\
c. Modify the Model and create Derivatives based on the Model, provided that such modifications and Derivatives are used exclusively for research and educational purposes. The Licensee may conduct experiments, perform analyses, and apply custom modifications to the Model to explore its capabilities and performance under various scenarios. If the Model is modified, the modified Model must include “EXAONE” at the beginning of its name. \\
\\
d. Distribute the Model and Derivatives in each case with a copy of this Agreement. \\
\newpage
\textbf{2.2 Scope of License:} The license granted herein does not authorize the Licensee to use the Model for any purpose not explicitly permitted under this Agreement. Any use beyond the scope of this license, including any commercial application or external distribution, is strictly prohibited unless explicitly agreed upon in writing by the Licensor. \\
\\
\\
\textbf{3. Restrictions}\label{textbf:Restrictions} \\
\\
\textbf{3.1 Commercial Use:} The Licensee is expressly prohibited from using the Model, Derivatives, or Output for any commercial purposes, including but not limited to, developing or deploying products, services, or applications that generate revenue, whether directly or indirectly. Any commercial exploitation of the Model or its derivatives requires a separate commercial license agreement with the Licensor. Furthermore, the Licensee shall not use the Model, Derivatives or Output to develop or improve any models that compete with the Licensor’s models. \\
\\
\textbf{3.2 Reverse Engineering:} The Licensee shall not decompile, disassemble, reverse engineer, or attempt to derive the source code, underlying ideas, algorithms, or structure of the Model, except to the extent that such activities are expressly permitted by applicable law. Any attempt to bypass or circumvent technological protection measures applied to the Model is strictly prohibited. \\
\\
\textbf{3.3 Unlawful Use:} The Licensee shall not use the Model and Derivatives for any illegal, fraudulent, or unauthorized activities, nor for any purpose that violates applicable laws or regulations. This includes but is not limited to the creation, distribution, or dissemination of malicious, deceptive, or unlawful content. \\
\\
\textbf{3.4 Ethical Use:} The Licensee shall ensure that the Model or Derivatives is used in an ethical and responsible manner, adhering to the following guidelines: \\
\\
a. The Model and Derivatives shall not be used to generate, propagate, or amplify false, misleading, or harmful information, including fake news, misinformation, or disinformation. \\
\\
b. The Model and Derivatives shall not be employed to create, distribute, or promote content that is discriminatory, harassing, defamatory, abusive, or otherwise offensive to individuals or groups based on race, gender, sexual orientation, religion, nationality, or other protected characteristics. \\
\\
c. The Model and Derivatives shall not infringe on the rights of others, including intellectual property rights, privacy rights, or any other rights recognized by law. The Licensee shall obtain all necessary permissions and consents before using the Model and Derivatives in a manner that may impact the rights of third parties. \\
\\
d. The Model and Derivatives shall not be used in a way that causes harm, whether physical, mental, emotional, or financial, to individuals, organizations, or communities. The Licensee shall take all reasonable measures to prevent misuse or abuse of the Model and Derivatives that could result in harm or injury. \\
\\
\\
\textbf{4. Ownership} \\
\\
\textbf{4.1 Intellectual Property:} All rights, title, and interest in and to the Model, including any modifications, Derivatives, and associated documentation, are and shall remain the exclusive property of the Licensor. The Licensee acknowledges that this Agreement does not transfer any ownership rights to the Licensee. All trademarks, service marks, and logos associated with the Model are the property of the Licensor. \\
\\
\textbf{4.2 Output:} Licensor claims no rights in Output. Licensee is solely responsible for the Output and its use. \\
\\
\textbf{4.3 Attribution:} In any publication or presentation of results obtained using the Model, the Licensee shall provide appropriate attribution to the Licensor, citing the Model's name and version, along with any relevant documentation or references specified by the Licensor. \\
\newpage
\textbf{5. No Warranty} \\
\\
\textbf{5.1 “As-Is” Basis:} The Model, Derivatives, and Output are provided on an “as-is” and “as-available” basis, without any warranties or representations of any kind, whether express, implied, or statutory. The Licensor disclaims all warranties, including but not limited to, implied warranties of merchantability, fitness for a particular purpose, accuracy, reliability, non-infringement, or any warranty arising from the course of dealing or usage of trade. \\
\\
\textbf{5.2 Performance and Reliability:} The Licensor does not warrant or guarantee that the Model, Derivatives or Output will meet the Licensee’s requirements, that the operation of the Model, Derivatives or Output will be uninterrupted or error-free, or that defects in the Model will be corrected. The Licensee acknowledges that the use of the Model, Derivatives or Output is at its own risk and that the Model, Derivatives or Output may contain bugs, errors, or other limitations. \\
\\
\textbf{5.3 No Endorsement:} The Licensor does not endorse, approve, or certify any results, conclusions, or recommendations derived from the use of the Model. The Licensee is solely responsible for evaluating the accuracy, reliability, and suitability of the Model for its intended purposes. \\
\\
\\
\textbf{6. Limitation of Liability} \\
\\
\textbf{6.1 No Liability for Damages:} To the fullest extent permitted by applicable law, in no event shall the Licensor be liable for any special, incidental, indirect, consequential, exemplary, or punitive damages, including but not limited to, damages for loss of business profits, business interruption, loss of business information, loss of data, or any other pecuniary or non-pecuniary loss arising out of or in connection with the use or inability to use the Model, Derivatives or any Output, even if the Licensor has been advised of the possibility of such damages. \\
\\
\textbf{6.2 Indemnification:} The Licensee agrees to indemnify, defend, and hold harmless the Licensor, its affiliates, officers, directors, employees, and agents from and against any claims, liabilities, damages, losses, costs, or expenses (including reasonable attorneys' fees) arising out of or related to the Licensee's use of the Model, any Derivatives, or any Output, including any violation of this Agreement or applicable laws. \\
\\
\\
\textbf{7. Termination} \\
\\
\textbf{7.1 Termination by Licensor:} The Licensor reserves the right to terminate this Agreement and revoke the Licensee’s rights to use the Model at any time, with or without cause, and without prior notice if the Licensee breaches any of the terms or conditions of this Agreement. Termination shall be effective immediately upon notice. \\
\\
\textbf{7.2 Effect of Termination:} Upon termination of this Agreement, the Licensee must immediately cease all use of the Model and Derivatives and destroy all copies of the Model and Derivatives in its possession or control, including any backup or archival copies. The Licensee shall certify in writing to the Licensor that such destruction has been completed. \\
\\
\textbf{7.3 Survival:} The provisions of this Agreement that by their nature should survive termination, including but not limited to, Sections 4 (Ownership), 5 (No Warranty), 6 (Limitation of Liability), and this Section 7 (Termination), shall continue to apply after termination. \\
\\
\\
\textbf{8. Governing Law} \\
\\
\textbf{8.1 Governing Law:} This Agreement shall be governed by and construed in accordance with the laws of the Republic of Korea, without regard to its conflict of laws principles. \\
\\
\textbf{8.2 Arbitration:} Any disputes, controversies, or claims arising out of or relating to this Agreement, including its existence, validity, interpretation, performance, breach, or termination, shall be referred to and finally resolved by arbitration administered by the Korean Commercial Arbitration Board (KCAB) in accordance with the International Arbitration Rules of the Korean Commercial Arbitration Board in force at the time of the commencement of the arbitration. The seat of arbitration shall be Seoul, Republic of Korea. The tribunal shall consist of one arbitrator. The language of the arbitration shall be English. \\
\\
\textbf{9. Alterations} \\
\\
\textbf{9.1 Modifications:} The Licensor reserves the right to modify or amend this Agreement at any time, in its sole discretion. Any modifications will be effective upon posting the updated Agreement on the Licensor’s website or through other means of communication. The Licensee is responsible for reviewing the Agreement periodically for changes. Continued use of the Model after any modifications have been made constitutes acceptance of the revised Agreement. \\
\\
\textbf{9.2 Entire Agreement:} This Agreement constitutes the entire agreement between the Licensee and Licensor concerning the subject matter hereof and supersedes all prior or contemporaneous oral or written agreements, representations, or understandings. Any terms or conditions of any purchase order or other document submitted by the Licensee in connection with the Model that are in addition to, different from, or inconsistent with the terms and conditions of this Agreement are not binding on the Licensor and are void. \\
\\
By downloading, installing, or using the EXAONE AI Model, the Licensee acknowledges that it has read, understood, and agrees to be bound by the terms and conditions of this Agreement. \\

\newpage
\section{Baseline models}
The models being compared are categorized into open-weight models: Small-size  models under 3B, Mid-size models between 10B and 30B, and Frontier models above 200B. Additionally, the models are divided into three types for performance evaluation, with specific details provided in Table~\ref{tab:baselines}.
\begin{table}[h!]
    \small
    \centering
    \setlength{\doublerulesep}{1pt}
    \begin{tabular}{m{1.8cm}|m{5cm}|wc{2cm}|wc{2.5cm}|wc{2cm}}
        \toprule
        \textbf{Category} & \textbf{Model} & \textbf{Parameters} & \textbf{Type} & \textbf{Release Date} \\
        \midrule
        Frontier & \href{https://huggingface.co/deepseek-ai/DeepSeek-R1-0528}{DeepSeek R1-0528}~\citep{deepseekai2025deepseekr1incentivizingreasoningcapability} & 671B (MoE) & Reasoning & May 2025 \\
        & \href{https://huggingface.co/deepseek-ai/DeepSeek-V3-0324}{DeepSeek V3-0324}~\citep{deepseekai2025deepseekv3technicalreport} & 671B (MoE) & Non-reasoning & Mar. 2025 \\
        & \href{https://huggingface.co/meta-llama/Llama-4-Maverick-17B-128E-Instruct}{Llama 4 Maverick} & 402B (MoE) & Non-reasoning & Apr. 2025 \\
        & \href{https://huggingface.co/Qwen/Qwen3-235B-A22B}{Qwen 3 235B}~\citep{yang2025qwen3technicalreport} & 235B (MoE) & Hybrid & Apr. 2025 \\
        \midrule
        Mid-size & \href{https://huggingface.co/Qwen/Qwen3-32B}{Qwen 3 32B}~\citep{yang2025qwen3technicalreport} & 32.8B & Hybrid & Apr. 2025 \\
        \rowcolor[rgb]{0.9,0.9,0.9} & \href{https://huggingface.co/LGAI-EXAONE/EXAONE-4.0-32B}{EXAONE 4.0 32B} & 32.0B & Hybrid & Jul. 2025 \\
        & \href{https://huggingface.co/google/gemma-3-27b-it}{Gemma 3 27B}~\citep{gemmateam2025gemma3technicalreport} & 27.4B & Non-reasoning & Mar. 2025 \\
        & \href{https://huggingface.co/mistralai/Mistral-Small-3.2-24B-Instruct-2506}{Mistral-Small-3.2-24B-Instruct-2506} & 24.0B & Non-reasoning & Jun. 2025 \\
        & \href{https://huggingface.co/mistralai/Magistral-Small-2506}{Magistral-Small-2506}~\citep{mistralai2025magistral} & 23.6B & Reasoning & Jun. 2025 \\
        & \href{https://huggingface.co/microsoft/Phi-4-reasoning-plus}{Phi 4 reasoning plus}~\citep{abdin2025phi4reasoningtechnicalreport} & 14.7B & Reasoning & Apr. 2025 \\
        & \href{https://huggingface.co/microsoft/phi-4}{Phi 4}~\citep{abdin2024phi4technicalreport} & 14.7B & Non-reasoning & Dec. 2024 \\
        \midrule
        Small-size & \href{https://huggingface.co/HuggingFaceTB/SmolLM3-3B}{SmolLM 3 3B} & 3.08B & Hybrid & Jul. 2025 \\
        & \href{https://huggingface.co/LGAI-EXAONE/EXAONE-Deep-2.4B}{EXAONE Deep 2.4B}~\citep{research2025exaonedeepreasoningenhanced} & 2.41B & Reasoning & Mar. 2025 \\
        & \href{https://huggingface.co/Qwen/Qwen3-1.7B}{Qwen 3 1.7B}~\citep{yang2025qwen3technicalreport} & 1.72B & Hybrid & Apr. 2025 \\
        \rowcolor[rgb]{0.9,0.9,0.9} & \href{https://huggingface.co/LGAI-EXAONE/EXAONE-4.0-1.2B}{EXAONE 4.0 1.2B} & 1.28B & Hybrid & Jul. 2025 \\
        & \href{https://huggingface.co/google/gemma-3-1b-it}{Gemma 3 1B}~\citep{gemmateam2025gemma3technicalreport} & 1.00B & Non-reasoning & Mar. 2025 \\
        & \href{https://huggingface.co/Qwen/Qwen3-0.6B}{Qwen 3 0.6B}~\citep{yang2025qwen3technicalreport} & 596M & Hybrid & Apr. 2025 \\
        \bottomrule
    \end{tabular}
    \vspace{2mm}
    \caption{The list of EXAONE 4.0 models and baseline models used for the evaluation along with their parameter size, type, and released date.}
    \label{tab:baselines}
\end{table}
\label{appendix:baseline_model}

\newpage
\section{Evaluation Details}
\label{appendix:evaluation_details}

\subsection{HELMET}
We include the \textsc{HELMET} benchmark~\citep{yen2025helmetevaluatelongcontextlanguage} in our evaluation to systematically assess models' long-context capabilities across both synthetic and real-world tasks. \textsc{HELMET} is designed as a comprehensive suite of diverse, application-centric tasks and addresses key limitations of prior benchmarks, such as inadequate input lengths, over-reliance on retrieval-style setups, and unreliable evaluation metrics. Crucially, it covers a wide spectrum of long-context challenges, including information recall, multi-hop retrieval, in-context generalization, and long-input generation, making it a well-suited benchmark for evaluating models' ability to process and reason over extended sequences in practical settings.

We adopt six of the seven categories from \textsc{HELMET}, \textit{Synthetic Recall}, \textit{Retrieval-Augmented Generation (RAG)}, \textit{Passage Re-ranking}, \textit{In-Context Learning (ICL)}, \textit{Long-document Question Answering (LongQA)}, and \textit{Summarization (Summ)}, to provide a balanced and holistic evaluation of long-context understanding and reasoning abilities.

\begin{figure}[h!]
    \centering
    \includegraphics[width=\textwidth]{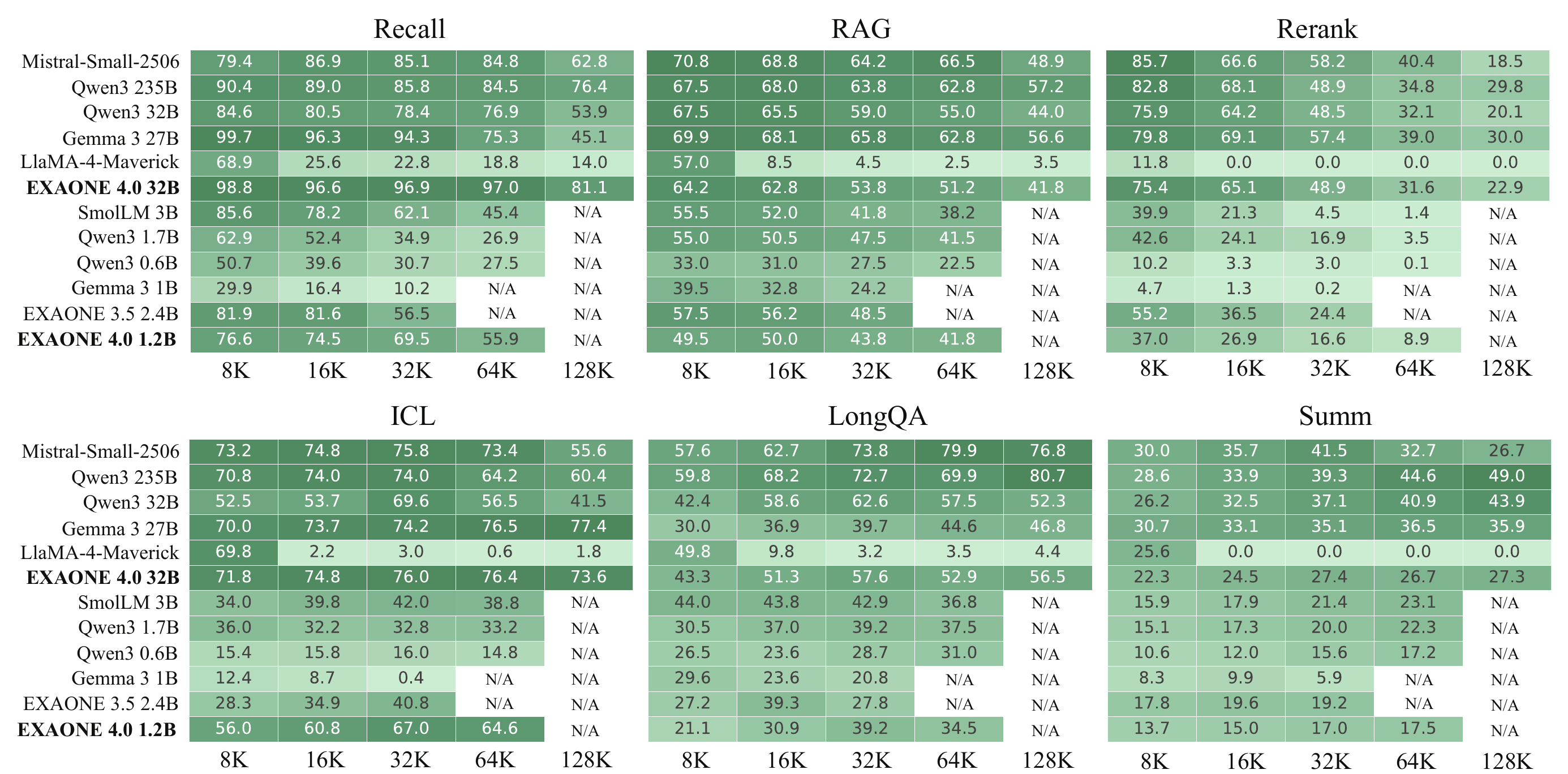}
    \caption{Performance of various models across six \textsc{HELMET} task categories, \textit{Recall}, \textit{RAG}, \textit{Passage Re-ranking}, \textit{ICL}, \textit{LongQA}, and \textit{Summarization}, at different context lengths (8K to 128K tokens). Darker cells indicate higher accuracy. Missing entries (\texttt{N/A}) denote models that do not support the corresponding input length or task.}
    \label{fig:helmet_figure}
\end{figure}

We formalize our decision to exclude \textit{LongCite} task from HELMET along three lines:
\begin{itemize}
\item Scope misalignment: HELMET emphasizes general long-context abilities, such as summarization, question answering, retrieval, and reasoning, whereas \textit{LongCite} centers on sentence-level citation accuracy, which constitutes a distinct attribution task rather than a core comprehension or generative skill.
\item Metric incompatibility: The benchmark employs standardized metrics like SubEM and model-based scoring, while \textit{LongCite} introduces specialized citation-precision and F1 measures. Integrating these heterogeneous metrics would compromise the uniformity essential for fair model comparison.
\item Benchmark coherence: Including a specialized citation task would divert HELMET from its unified objective of comparing long-context reasoning across models. Such an inclusion would introduce extraneous variability and diminish comparative consistency.
\end{itemize}
Consequently, omitting \textit{LongCite} ensures that HELMET remains a concise, cohesive benchmark focused solely on evaluating long-context language modeling capabilities across properly aligned tasks. Detailed task-wise scores are reported in Table~\ref{tab:helmet_per_tasks}, while Figure~\ref{fig:helmet_figure} illustrates how performance on each task varies across different context lengths.

\begin{table*}[h]
\centering
\renewcommand{\arraystretch}{1.1}
\small
\begin{tabular}{c|l|c|cccccc}
\toprule
\multicolumn{1}{c|}{\textbf{Context Len.}} & 
\multicolumn{1}{c|}{\textbf{Model}} &
\multicolumn{1}{c|}{\textbf{Total Avg.}} & 
\textbf{Recall} & 
\textbf{RAG} & 
\textbf{LongQA} & 
\textbf{Summ} & 
\textbf{Rerank} & 
\textbf{ICL} \\
\midrule
\rowcolor[rgb]{0.9,0.9,0.9}\multicolumn{9}{c}{\textsc{Mid-Size}} \\
\midrule
\multirow{5}{*}{128K}
  & Mistral-Small-2506   & 61.93 & 79.82 & 63.83 & 70.17 & 33.32 & 53.87 & 70.56 \\
  & Qwen3 235B           & 63.33 & 85.23 & 63.85 & 70.26 & 39.08 & 52.89 & 68.68 \\
  & Qwen3 32B            & 54.47 & 74.88 & 58.20 & 54.67 & 36.12 & 48.17 & 54.78 \\
  & Gemma 3 27B         & 58.34 &  82.16 & 64.63 & 39.59 & 34.26 & 55.06  & 74.38 \\
  & LlaMA-4-Maverick     & 13.72 & 30.00 & 15.20 & 14.14 & 5.12  & 2.36  & 15.48 \\
  & EXAONE 4.0 32B       & 58.34 & 94.06 & 54.75 & 52.31 & 25.64 & 48.78 & 74.52 \\
\midrule
\rowcolor[rgb]{0.9,0.9,0.9}\multicolumn{9}{c}{\textsc{Small-Size}} \\
\midrule
\multirow{6}{*}{32K}
  & SmolLM 3B        & 41.25 & 75.29 & 49.75 & 43.55 & 18.40 & 21.92 & 38.60 \\
  & Qwen3 1.7B       & 35.94 & 50.07 & 51.00 & 35.56 & 17.47 & 27.89 & 33.67 \\
  & Qwen3 0.6B       & 21.85 & 40.33 & 30.50 & 26.28 & 12.73 & 5.51  & 15.73 \\
  & Gemma 3 1B       & 15.49 & 18.86 & 32.17 & 24.70 & 8.03  & 2.08  & 7.13 \\
  & EXAONE 3.5 2.4B  & 41.85 & 73.35 & 54.08 & 31.44 & 18.87 & 38.69 & 34.68 \\
  & EXAONE 4.0 1.2B  & 42.50 & 73.52 & 47.75 & 30.43 & 15.23 & 26.80 & 61.27 \\
\midrule
\multirow{6}{*}{64K}
  & SmolLM 3B        & 38.60 & 67.81 & 46.88 & 41.86 & 19.58 & 16.80 & 38.65 \\
  & Qwen3 1.7B\textsuperscript{†}       & 33.83 & 44.29 & 48.63 & 36.04 & 18.68 & 21.80 & 33.55 \\
  & Qwen3 0.6B\textsuperscript{†}       & 21.10 & 37.13 & 28.50 & 27.47 & 13.85 & 4.15  & 15.50 \\
  & Gemma 3 1B       & N/A   & N/A   & N/A   & N/A   & N/A   & N/A   & N/A \\
  & EXAONE 3.5 2.4B  & N/A   & N/A   & N/A   & N/A   & N/A   & N/A   & N/A \\
  & EXAONE 4.0 1.2B  & 41.17 & 69.13 & 46.25 & 31.44 & 15.80 & 22.33 & 62.10 \\
\bottomrule
\end{tabular}
\caption{Comparison of \textsc{Mid-Size} and \textsc{Small-Size} models across tasks on the \textsc{HELMET} benchmark.
(\texttt{N/A}) indicates that models not supporting specific input lengths are omitted from evaluation. 
(\textsuperscript{†}) denotes that models are extended to 64K context length using YaRN.
}
\label{tab:helmet_per_tasks}
\end{table*}

\subsection{RULER}
We evaluate our model's long-context capabilities using the RULER benchmark~\citep{hsieh2024rulerwhatsrealcontext}, a synthetic evaluation suite designed to assess various aspects of long-context understanding beyond simple retrieval. RULER consists of diverse task categories, including retrieval, multi-hop tracing, aggregation, and question answering, and supports flexible configurations for context length and task complexity. The performance of our models across different sequence lengths is summarized in Table~\ref{tab:ruler_results}.

\begin{table*}[h]
\centering
\renewcommand{\arraystretch}{1.2}
\small
\begin{tabular}{l|cccccc}
\toprule
\multicolumn{1}{c|}{\textbf{Model}} & 
\textbf{4K} & 
\textbf{8K} & 
\textbf{16K} & 
\textbf{32K} & 
\textbf{64K} & 
\textbf{128K} \\
\midrule
\rowcolor[rgb]{0.9,0.9,0.9}\multicolumn{7}{c}{\textsc{Mid-Size}} \\
\midrule
Mistral-Small-2506               & 97.18 & 97.15 & 96.66 & 94.57 & 88.53 & 71.84 \\
Qwen3 235B                       & 97.70 & 97.20 & 96.40 & 95.10 & 93.30 & 90.60 \\
Qwen3 32B                        & 98.40 & 96.00 & 96.20 & 94.40 & 91.80 & 85.60 \\
Geamma3 27B                    & 95.5 & 95.13 & 93.88 & 91.1  & 80.59  & 66.00 \\
LlaMA-4-Maverick                 & 97.10 & 96.85 & 12.96 & 4.90  & 4.35  & 2.85 \\
EXAONE 4.0 32B                   & 96.26 & 94.85 & 93.93 & 93.64 & 91.73 & 88.18 \\

\midrule
\rowcolor[rgb]{0.9,0.9,0.9}\multicolumn{7}{c}{\textsc{Small-Size}} \\
\midrule
SmolLM 3B                        & 92.30 & 85.01 & 81.76 & 77.85 & 66.27 & - \\
Qwen3 1.7B                       & 89.70 & 86.58 & 80.23 & 75.17 & 65.94\textsuperscript{†} & - \\
Qwen3 0.6B                       & 80.74 & 73.64 & 67.17 & 60.82 & 55.09\textsuperscript{†} & - \\
Gemma 3 1B                       & 58.93 & 46.98 & 41.09 & 28.75 & N/A    & - \\
EXAONE 3.5 2.4B                  & 88.91 & 87.79 & 87.27 & 77.73 & N/A     & - \\
EXAONE 4.0 1.2B                  & 87.02 & 86.71 & 88.83 & 81.07 & 77.43 & - \\
\bottomrule
\end{tabular}
\caption{Accuracy scores of \textsc{Mid-Size} and \textsc{Small-Size} models on the \textsc{RULER} benchmark across varying context lengths (4K to 128K tokens). 
(\texttt{N/A}) indicates that models not supporting specific input lengths are omitted from evaluation. 
(\textsuperscript{†}) denotes that models are extended to 64K context length using YaRN.
}
\label{tab:ruler_results}
\end{table*}

\subsection{LongBench}
LongBench~\citep{bai2024longbenchbilingualmultitaskbenchmark} has been suggested as a bilingual benchmark to assess long context comprehension in English and Chinese.
We focus on the English subsets, specifically \textit{Single-doc QA}, \textit{Multi-doc QA}, \textit{Summarization}, and \textit{Few-shot Learning}. 

The \textit{Single-doc QA} task covers datasets such as NarrativeQA~\citep{kovcisky2018narrativeqa}, Qasper~\citep{dasigi-etal-2021-dataset}, and MultiFieldQA-EN~\citep{bai2024longbenchbilingualmultitaskbenchmark}. For the \textit{Multi-doc QA} task, we employ benchmarks including HotpotQA~\citep{yang-etal-2018-hotpotqa}, 2WikiMultihopQA~\citep{ho-etal-2020-constructing}, and MuSiQue~\citep{trivedi-etal-2022-musique}. The \textit{Summarization} task utilizes datasets like GovReport~\citep{huang-etal-2021-efficient}, QMSum~\citep{zhong-etal-2021-qmsum}, and MultiNews~\citep{fabbri-etal-2019-multi}. For the \textit{Few-shot Learning} task, we draw from TREC~\citep{li-roth-2002-learning} and TriviaQA~\citep{joshi-etal-2017-triviaqa}. All evaluations follow the official protocols and metrics defined in LongBench.

Comprehensive results for each task are shown in Table~\ref{tab:longbench_details}.

\begin{table*}[h]
\centering
\renewcommand{\arraystretch}{1.2}
\small
\begin{tabular}{c|l|c|cccc}
\toprule
\multicolumn{1}{c|}{\textbf{Context Len.}} & 
\multicolumn{1}{c|}{\textbf{Model}} &
\multicolumn{1}{c|}{\textbf{Total Avg.}} &
\textbf{Single-doc QA} & 
\textbf{Multi-doc QA} & 
\textbf{Summarization} & 
\textbf{Few-shot Learning}
 \\
\midrule
\rowcolor[rgb]{0.9,0.9,0.9}\multicolumn{7}{c}{\textsc{Mid-Size}} \\
\midrule
\multirow{5}{*}{128K}
  & Mistral-Small-2506   & 51.48 & 43.73 & 52.51 & 28.82 & 80.87 \\
  & Qwen3 235B           & 45.28 & 41.45 & 46.96 & 25.56 & 67.13 \\
  & Qwen3 32B            & 44.24 & 41.27 & 47.97 & 25.73 & 62.01 \\
  & Gemma3 27B           & 51.54 & 42.65 & 54.81 & 24.45 & 84.26 \\
  & LlaMA-4-Maverick     & 34.71 & 32.72 & 24.68 & 23.84 & 57.58 \\
  & EXAONE 4.0 32B       & 48.12 & 39.40 & 48.46 & 27.34 & 77.28 \\

\midrule
\rowcolor[rgb]{0.9,0.9,0.9}\multicolumn{7}{c}{\textsc{Small-Size}} \\
\midrule
\multirow{6}{*}{32K}
  & SmolLM 3B        & 39.85 & 33.38 & 18.26 & 27.94 & 79.83 \\
  & Qwen3 1.7B       & 41.82 & 33.61 & 31.87 & 26.16 & 75.63 \\
  & Qwen3 0.6B       & 32.72 & 22.75 & 20.29 & 23.11 & 64.73 \\
  & Gemma 3 1B       & 34.91 & 24.85 & 24.09 & 21.41 & 69.29 \\
  & EXAONE 3.5 2.4B  & 42.74 & 35.03 & 43.11 & 20.05 & 72.75 \\
  & EXAONE 4.0 1.2B  & 36.75 & 30.93 & 34.98 & 25.14 & 55.96 \\
\midrule
\multirow{6}{*}{64K}
  & SmolLM 3B        & 39.93 & 33.53 & 18.27 & 28.11 & 79.83 \\
  & Qwen3 1.7B\textsuperscript{†}       & 41.92 & 32.01 & 32.53 & 25.95 & 77.19 \\
  & Qwen3 0.6B\textsuperscript{†}       & 32.44 & 22.38 & 21.40 & 23.15 & 62.84 \\
  & Gemma 3 1B       & N/A   & N/A   & N/A   & N/A   & N/A \\
  & EXAONE 3.5 2.4B  & N/A   & N/A   & N/A   & N/A   & N/A \\
  & EXAONE 4.0 1.2B  & 36.93 & 31.02 & 35.09 & 25.28 & 56.33 \\
\bottomrule
\end{tabular}
\caption{
Task-wise performance of \textsc{Mid-Size} and \textsc{Small-Size} models on the LongBench benchmark across four task categories: \textit{Single-doc QA}, \textit{Multi-doc QA}, \textit{Summarization}, and \textit{Few-shot Learning}. 
Each score represents the average accuracy over the English subset of LongBench at specified context lengths. 
(\texttt{N/A}) indicates that models not supporting specific input lengths are omitted from evaluation. 
(\textsuperscript{†}) denotes that models are extended to 64K context length using YaRN.
}
\label{tab:longbench_details}
\end{table*}

\clearpage
\subsection{Ko-LongBench}
\label{appendix:ko_longbench}

Ko-LongBench is an in-house benchmark developed to evaluate long-context understanding in Korean. It consists of multiple tasks, including \textit{Document QA}, \textit{Story Understanding}, \textit{Dialogue History Understanding}, \textit{In-Context Learning}, \textit{Structured QA}, and \textit{RAG}, allowing for a comprehensive assessment of LLMs' long-context capabilities in real-world scenarios. A detailed overview of the dataset is provided in Table~\ref{tab:ko_long_dis}, and representative prompt examples for each task are shown in Figures~\ref{fig:ko_longbench} and~\ref{fig:ko_longbench2}. Table~\ref{tab:ko_longbench_avg} summarizes the average performance of \textsc{Small-Size} models on Ko-LongBench, reporting scores for both 32K and 64K context lengths.

\begin{table}[h]
    
    \centering\resizebox{\textwidth}{!}{
    \renewcommand{\arraystretch}{1.2}
    \begin{tabular}{l|c|c|l}
        \toprule
        \textbf{Category} & \textbf{Subtask} & \textbf{\# Samples} & \textbf{Description}  \\
        \midrule
        {\multirow{4}{*}{\makecell[l]{SingledocQA /\\ MultidocQA}}} & Medical & 300 & Single- and Multi-Document Question Answering in the Medical Domain \\
         & Legal & 300 & Single- and Multi-Document Question Answering in the  Legal Domain \\
         & Finance & 300 & Single- and Multi-Document Question Answering in the Finance Domain \\
         & Patent & 300 & Single- and Multi-Document Question Answering in the Patent Domain \\
         
        \midrule
        {\multirow{2}{*}{Story Understanding}} & Ordering & 66 & Evaluation of the Ability to Sequence the Given Story \\
        & Mixeing & 150 & Evaluation of the Ability to Infer the number of Mixed Stories \\
        
        \midrule
        {\multirow{3}{*}{\makecell[l]{Long-dialogue \\History Understanding}}} & Wrong chatbot & 150 & Inferring Inconsistencies with Given Information in Multi-turn Dialogues \\
        & Wrong inference & 150 & Inferring Information that cannot be deduced from Multi-turn Dialogues \\
        & Topic classify & 150 & Evaluating the Ability to classify topics in Multi-turn Dialogues \\
        
        \midrule
        {\multirow{2}{*}{\makecell[l]{Long In-context \\Learning}}} & Manual QA & 150 & Evaluation of Information Extraction Ability based on Product Manuals \\
        & Many-Shot & 150 & Evaluation of Information Extraction ability within a Few-shot Context \\

        \midrule
        Long Structued QA & Table QA & 300 & Table-Based Question Answering : Evaluation of Table Interpretation Skills \\

        \midrule
        {\multirow{2}{*}{RAG}} & Manual QA & 150 & Single-Document Question on Retrieved Document Context \\
        & MultiQA & 150 & Multi-Document Question based on Retrieved Document Context \\
        
        \midrule
        \textbf{Total} &   & \textbf{2766} & \\
        
        \bottomrule
    \end{tabular}}
    \vspace{2mm}    
    \caption{Descriptions of Ko-LongBench.}
    \label{tab:ko_long_dis}
    
\end{table}

\begin{figure}[H]
\centering
\begin{tcolorbox}[
  title=Ko-LongBench Example (Long-dialogue History Understanding),
  colframe=Black!80!White,
  colback=gray!10,
  coltitle=white,
  colbacktitle=Black!80!White,
  fonttitle=\bfseries,
  breakable=false,
  rounded corners,
  boxsep=3pt,
  width=\textwidth
]

다음 문제에 대해 정답을 고르세요. 당신의 최종 정답은 ABCD 중 하나이고, 정답: 뒤에 와야 합니다. 정답을 고르기 전에 차근차근 생각하고 추론하세요.\\\\
 
[Dialogue 0]\\안녕하세요. 50대 남성입니다. 반갑습니다. 반갑습니다 저는 20대 여성입니다 그러시군요~ 혹시 거주하는 곳이 어디인가요? 저는 경상도에 거주하고 있어요~ 선생님은요? 저는 경기도에 거주하고 있어요~ 혹시 직업이 있으신가요? 저는 아직까지는 학생입니다. 선생님은 있으신가요? 저는 그냥 일반 직장인이랍니다. 혹시 선생님은 시험기간에 밤을 새시나요? 자주 새요.. 미리 해야하는데 …\textsl{<중략>}\\

질문 : 위 대화들이 주로 다루고 있는 메인 토픽은 무엇인가?\\
A) 미용과 건강>건강
B) 주거와 생활
C) 개인 및 관계>연애/결혼
D) 여가와 오락>게임\\

정답: \{\textsl{answer}\}\\

\end{tcolorbox}
\caption{Example of Long-dialogue History Understanding (Topic classification) in Ko-LongBench.}
\label{fig:ko_longbench}
\end{figure}

\begin{figure}[H]
\centering
\begin{tcolorbox}[
  title=Ko-LongBench Example (Long Structued QA),
  colframe=Black!80!White,
  colback=gray!10,
  coltitle=white,
  colbacktitle=Black!80!White,
  fonttitle=\bfseries,
  breakable=false,
  rounded corners,
  boxsep=3pt,
  width=\textwidth
]

다음 문제에 대해 정답을 고르세요. 당신의 최종 정답은 ABCD 중 하나이고, 정답: 뒤에 와야 합니다. 정답을 고르기 전에 차근차근 생각하고 추론하세요.\\\\
 
문서 1:인천봉화초등학교 다목적교실 조성공사 시방서(건축) <table><tbody><tr><td>바탕의 종류</td><td>도장 종류</td><td>공법</td></tr><tr><td rowspan='3'>목재면, 플라스터면, 모르타르면, 콘크리트면</td><td>1종</td><td> …\textsl{<중략>}\\

질문 : 2022년 경기도청 북부청사 소방시설 점검 및 소방 안전관리 대행 용역에서 본관과 별관의 면적 합계는 전체 면적의 약 몇 퍼센트를 차지하는가?\\
A) 약 50\%
B) 약 60\%
C) 약 70\%
D) 약 80\%\\

정답: \{\textsl{answer}\}\\

\end{tcolorbox}
\caption{Example of Long Structued QA (Table QA) in Ko-LongBench.}
\label{fig:ko_longbench2}
\end{figure}

\begin{table*}[h]
\centering
\renewcommand{\arraystretch}{1.2}
\small
\begin{tabular}{l|cc}
\toprule
\multicolumn{1}{c|}{\textbf{Model}} &
\textbf{Avg. up to 32K} & \textbf{Avg. up to 64K} \\
\midrule
SmolLM 3B         & 19.3 & 15.7 \\
Qwen3 1.7B        & 62.4 & 57.1\textsuperscript{†} \\
Qwen3 0.6B        & 18.6 & 16.4\textsuperscript{†} \\
Gemma 3 1B        & 6.3  & N/A \\
EXAONE 3.5 2.4B   & 57.8  & N/A \\
EXAONE 4.0 1.2B   & 72.0 & 69.8 \\
\bottomrule
\end{tabular}
\caption{
Average performance of \textsc{Small-Size} models on Ko-LongBench, a multi-task benchmark designed to evaluate long-context understanding in Korean. 
The left column reports the average scores across all tasks up to 32K context length, while the right column shows the average scores up to 64K.
(\texttt{N/A}) indicates that models not supporting specific input lengths are omitted from evaluation. 
(\textsuperscript{†}) denotes that models are extended to 64K context length using YaRN.
}
\label{tab:ko_longbench_avg}
\end{table*}

\subsection{\textsc{WMT24++}}
\label{appendix:wmt24pp}
Figure~\ref{fig:wmt_judge_prompt} presents the prompt used for LLL-as-a-judge in \textsc{WMT24++} benchmark. We use the same 0-shot prompt from the official \textsc{WMT24++} paper.

\begin{figure}[H]
\centering
\begin{tcolorbox}[
  title=WMT24++ Judge Prompt,
  colframe=Black!80!White,
  colback=gray!10,
  coltitle=white,
  colbacktitle=Black!80!White,
  fonttitle=\bfseries,
  breakable=false,
  rounded corners,
  boxsep=3pt,
  width=\textwidth
]

"You are a professional judge for evaluating the quality of \{src\_lang\} to \{tgt\_lang\} translations suitable for use in \{tgt\_region\}. Based on the source text, the human-written translation, and machine translation surrounded with triple backticks, your task is to assess the quality of the machine translation on a continuous scale from 0 to 100. A score of 0 means "No meaning preserved," then the scale goes through "Some meaning preserved," to "Most meaning preserved and few grammatical mistakes," up to a score of 100, which means "Perfect meaning and grammar." Your output should only include the score from 0 to 100 without any additional text.\\\\

\{src\_lang\} text: ```\{src\_text\}''' \\
\{tgt\_lang\} human translation: ```\{tgt\_text\}''' \\
\{tgt\_lang\} machine translation: ```\{model\_text\}'''\\

\end{tcolorbox}
\caption{The judge prompt for evaluating translation quality in \textsc{WMT24++} benchmark.}
\label{fig:wmt_judge_prompt}
\end{figure}

\newpage
\bibliographystyle{plain} 
\bibliography{refs} 


\end{document}